\documentclass[a4paper,fleqn]{cas-dc}
\usepackage[numbers]{natbib}
\usepackage{algorithm,algorithmic}
\usepackage{tikz}
\usepackage{caption}
\usetikzlibrary{positioning, arrows.meta}  
\usepackage{hyperref}

\usepackage{tabularx} 
\usepackage{booktabs} 
\usepackage{listings} 
\usepackage{xcolor} 
\usepackage{textcomp}
\def\tsc#1{\csdef{#1}{\textsc{\lowercase{#1}}\xspace}}
\tsc{WGM}
\tsc{QE}
\tsc{EP}
\tsc{PMS}
\tsc{BEC}
\tsc{DE}
\begin{document}
\let\WriteBookmarks\relax
\def\floatpagepagefraction{1}
\def\textpagefraction{.001}
\shorttitle{MIMIC Dataset}
\shortauthors{A. Khaled et~al.}
\title [mode = title]{Leveraging MIMIC Datasets for Better Digital Health: A Review on Open Problems, Progress Highlights, and Future Promises}    
\author[1]{Afifa Khaled}[
                        orcid=0000-0001-5500-9940]
\cormark[1]
\ead{afifakhaled@mail.ustc.edu.cn}
\affiliation[1]{organization={Suzhou Institute for Advanced Research, University of Science and Technology of China},
                  city={Suzhou},
                country={China}}

\author[1]{Mohammed Sabir}

\author[2]{Rizwan Qureshi}
\affiliation[2]{organization={Center for Research in Computer Vision, University of Central Florida},
                city={Florida},
                country={USA}}

\author[3]{Camillo Maria Caruso}
\affiliation[3]{organization={Research Unit for Computer Systems and Bioinformatics, Department of Engineering, Università Campus Bio-Medico di Roma},
                city={Rome},
                country={Italy}}

\author[3]{Valerio  Guarrasi}
\author[4]{Suncheng Xiang}
\affiliation[4]{organization={Shanghai Jiao Tong University},
                city={Shanghai},
                country={China}}
                
\author[1]{S Kevin Zhou}
\cormark[1]
\ead{skevinzhou@ustc.edu.cn}

\cortext[cor1]{Corresponding author}

\begin{abstract}
The Medical Information Mart for Intensive Care (MIMIC) datasets have become the Kernel of Digital Health Research by providing freely accessible, deidentified records from tens of thousands of critical care admissions, enabling a broad spectrum of applications in clinical decision support, outcome prediction, and healthcare analytics. Although numerous studies and surveys have explored the predictive power and clinical utility of MIMIC based models, critical challenges in data integration, representation, and interoperability remain underexplored. This paper presents a comprehensive survey that focuses uniquely on open problems. We identify persistent issues such as data granularity, cardinality limitations, heterogeneous coding schemes, and ethical constraints that hinder the generalizability and real-time implementation of machine learning models. We highlight key progress in dimensionality reduction, temporal modelling, causal inference, and privacy preserving analytics, while also outlining promising directions including hybrid modelling, federated learning, and standardized preprocessing pipelines. By critically examining these structural limitations and their implications, this survey offers actionable insights to guide the next generation of MIMIC powered digital health innovations.
\end{abstract}

\maketitle
\begin{figure*}[h]
    \centering
    \includegraphics[width=0.9\textwidth]{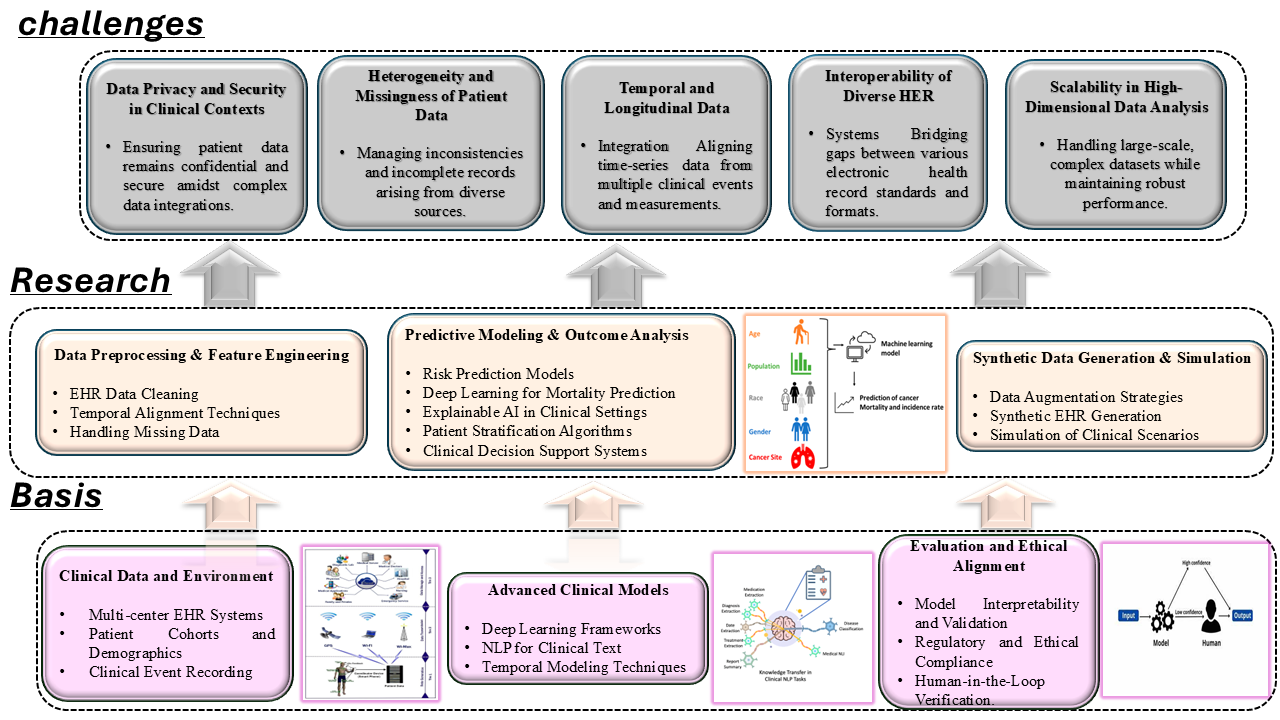} 
    \caption{The overall framework of our survey.}
    \label{Frameworrk}
\end{figure*}
\section{Overview}


Rapid advancement of digital health technologies has catalyzed a paradigm shift in how clinical data is leveraged to improve patient outcomes\cite{Johnson2023}, inform policy decisions\cite{Johnson2023}, and drive biomedical innovation\cite{Medhat2023}. Among the many data resources that fuel this transformation, the Medical Information Mart for Intensive Cares (MIMICs) datasets have emerged as the Kernel in critical care informatics, offering a unique combination of scale, granularity, and accessibility\cite{MITLab2020}. Figure~\ref{Frameworrk} shows the overall framework of our survey.

\subsection{MIMICs Datasets}

The Medical Information Mart for Intensive Care (MIMIC) dataset is a widely recognized resource for critical care research and healthcare improvement. Developed by the Laboratory for Computational Physiology (LCP) at the Massachusetts Institute of Technology (MIT) in collaboration with the Beth Israel Deaconess Medical Center (BIDMC) in Boston, USA~\cite{mimicintro}, MIMIC contains deidentified clinical data from over 40,000 patients in the intensive care unit (ICU). Initiated in the early 2000s, the project aimed to systematically collect and structure electronic health records (EHRs) of ICU patients, encompassing diverse data types such as physiological signals~\cite{qureshi2017multistage}, laboratory results~\cite{huang2017using}, medication records, and demographic information~\cite{feder2020active}. Over time, MIMIC has expanded to incorporate data from multiple facilities and cover a broader range of medical conditions \cite{mimicdatause}. The dataset has since become a cornerstone in research areas including disease classification, outcome prediction, and personalized treatment planning, supporting advancements in both machine learning applications and clinical decision-making \cite{Johnson2016mimic}. 
Later, MIMICs datasets have been developed by the MIT Laboratory for Computational Physiology, the MIMIC dataset series (MIMIC-II, III, IV, and extensions such as MIMIC-CXR and MIMIC-Waveform) encompasses rich, deidentified data from tens of thousands of admissions to intensive care units (ICU)\cite{MITLab2020}, including structured physiological measurements, unstructured clinical notes\cite{CERN2015}, time series waveforms, and medical imaging\cite{MITLab2020}. These resources enable researchers to address challenges in data interoperability \cite{Medhat2023} and advance evidence informed policy \cite{Johnson2023} through scalable analytical frameworks \cite{OECD2023}.
The MIMIC datasets have significantly advanced digital health research by offering open access to richly detailed, de-identified electronic health records (EHRs) from ICU patients. First published as MIMIC-III, the data set includes more than 40,000 stays in the ICU, featuring minute-level physiological data, laboratory results, medications, procedures, and clinical notes \cite{Johnson2016}. The 2020 release of MIMIC-IV expanded this to more than 65,000 ICU admissions and 200,000 emergency visits, with enhanced data schemas and standardized coding \cite{turn0search2}. 

 During the past decade, MIMIC has enabled a broad spectrum of research \cite{MITLab2020}, from mortality prediction and sepsis detection \cite{Johnson2023} to clinical natural language processing \cite{OECD2023} and epidemiological surveillance\cite{MITLab2020}. Its open access model has fostered reproducibility, standardized benchmarking, and interdisciplinary collaboration between clinical medicine, machine learning~\cite{qureshi2023artificial}, and public health\cite{Johnson2023,qureshi2023ai}. Despite their extensive use, the increasing complexity of modern AI methods \cite{Medhat2023} and emerging healthcare needs require a comprehensive reevaluation of how these datasets are used and what novel research directions they can support\cite{OECD2023}.

\textbf{MIMICs datasets comprises components that individually represent facets of the ICU experience for example of these informations:}
\begin{itemize}
    \item {\it {Patient Details:}} Includes data regarding the age range of patients and their gender distribution, along with any background information such as ethnicity. 
    \item {\it {Medical Records:}} Comprise, in-depth documentation of diagnoses, treatments, and comments from healthcare providers.
    \item {\it {Biological Indicators:}} Information such as pulse rate,  blood pressure measurements, and oxygen saturation levels recorded using monitoring tools. 
   \item  {\it {Laboratory Findings:}} Information regarding blood tests' outcomes and results from imaging studies and various lab measurements. 
   \item {\it {Medication Information:}} In-depth information about the medications given while patients were in the ICU.
    \item {\it {Data on Mortality and Results:}} This category covers information on survival rates and the results of treatments given while in the ICU. 
\end{itemize}

The information is carefully anonymized to safeguard confidentiality and enable researchers to access it in compliance with ethical guidelines like the Health Insurance Portability and Accountability Act (HIPAA)~\cite{gostin2009beyond}. It is accessible for research purposes at no cost; however researchers are required to sign a data use agreement (DUA) and complete a course on data privacy. The recent iteration of the MIMIC III datasets is currently available via the PhysioNet platform, which also houses other health related datasets \cite{mimicaccess}. 

Researchers have extensively utilized the MIMIC datasets to predict ICU mortality and complications, detect early warning signs of critical conditions, personalize treatments, evaluate care effectiveness, and develop AI-driven clinical decision support systems. These efforts underscore the transformative potential of integrating clinical datasets with computational methods to enhance patient care. MIMIC continues to serve as a pivotal resource in advancing healthcare research and promoting the adoption of data science in clinical practice \cite{mimicaccess}.

Previous reviews of MIMIC data sets (MIMIC-III and MIMIC-IV) have focused mainly on narrow domains, such as predictive modeling or natural language processing, without fully capturing the breadth of the evolving utility of MIMIC \cite{johnson2016data}. These surveys rarely explore the cross-domain potential of MIMIC data, such as its application in imaging, clinical decision support, and fairness-aware AI systems \cite{MIMICCXR2019}. 

In addition, emerging trends in AI in healthcare care, such as temporal learning (modeling how patient data evolve over time), multimodal fusion (combining structured data, notes and imaging), explainable AI (XAI), fairness and domain adaptation, have not been systematically reviewed in the context of MIMIC-based research \cite{johnson2023mimic}. There remains a critical need for a unified, up-to-date survey that bridges modern machine learning methodologies with the rich, multimodal content of MIMIC datasets (e.g. lab results, clinical notes and X-rays) \cite{BiomedGPT2024}. Such a survey is essential to connect clinical challenges (e.g., early disease detection, risk stratification) with cutting-edge AI techniques to advance precision healthcare.

\textbf{In this paper}, we address these gaps by providing a structured and comprehensive review of MIMIC-based research, categorizing previous work into two broad domains.
\begin{itemize}
    \item \textit{(1)} traditional clinical and computational applications, and
    \item \textit{(2)} novel and advanced data mining approaches.
\end{itemize}
We also present a unified taxonomy of use cases, identify persistent methodological challenges, and highlight emerging opportunities at the intersection of artificial intelligence and clinical relevance.

\subsection{Recent Advancements on  MIMICs Datasets}

MIMICs datasets are central to health informatics and machine learning research. Numerous surveys have examined their usage patterns, identified key challenges, and proposed improvements. Below is a brief overview of significant surveys exploring various aspects of MIMICs datasets utilization.

\subsubsection{Deep EHR: Advances in Deep Learning for Electronic Health Record Analysis}

One of the earliest comprehensive surveys, “Deep EHR,” systematically reviewed DL applications to EHR analysis, categorizing approaches into information extraction, representation learning, outcome prediction, phenotyping, and de identification .

Over the last twenty years, the exponential growth in the quantity and complexity of EHRs has created an urgent requirement for sophisticated representation learning techniques to extract clinically significant features and enhance predictive modeling activities\cite{shickel2017deep}. Deep learning (DL) methods\cite{turn0search1}, notably the authors highlighted that recurrent neural networks (RNNs) and long-short term memory (LSTM) architectures excel at capturing temporal dependencies in time series data, while autoencoders and variational methods provide compact latent embeddings that improve downstream classifier performance, feedforward neural networks, convolutional models, and recurrent networks, have proven to be highly effective in processing heterogeneous EHR information. These methods have outperformed traditional machine learning techniques in critical tasks such as predicting outcomes, identifying patient phenotypes, and modeling temporal data \cite{turn0search2}.

Subsequent evaluations built upon these foundational concepts by exploring sophisticated architectures and techniques for data augmentation. For example, Zhengping Che and his team developed  \textit{ehrGAN}, a generative adversarial network aimed at generating realistic electronic health record (EHR) sequences for semi-supervised risk prediction, resulting in notable improvements in prediction tasks of the onset \cite{turn0academia22}. In a similar vein, unsupervised representation learning through deep autoencoders and their denoising variants has demonstrated the ability to capture underlying patient phenotypes, thereby enhancing clustering methods and early warning systems \cite{10441762}.

Taxonomies developed in subsequent surveys have further elucidated our understanding of DL methods for EHR. A review of MDPI by Jiabao Xu \textit{ et al.} 2022 among other researchers classified these techniques into four main types: information extraction, representation learning, medical prediction, and privacy protection. This categorization was accompanied by a benchmarking of leading architectures across well known datasets such as MIMIC, eICU, and i2b2 \cite{turn0search9}.

Furthermore, another comparative study investigated key deep learning architectures, including convolutional neural networks (CNNs), RNNs, and Transformers. The analysis revealed that convolutional models are particularly adept at capturing local temporal patterns in irregularly sampled vital signs. Conversely, attention based Transformer models provide superior interpretability by emphasizing critical time points in the data\cite{AYALASOLARES2020103337}.

Advancements in taxonomy and architectural evaluation are crucial for improving deep learning (DL) applications in electronic health records (EHRs), enabling better alignment of techniques with clinical tasks. However, evaluations on datasets like MIMIC-III reveal challenges such as sparse and irregular data, where traditional RNNs underperform. This has led to the development of time-aware models like GRU-D, which address missing data through masking and decay mechanisms \cite{turn0search14}. Furthermore, DL models have shown promise in improving early disease detection and personalized screening using longitudinal diagnostic and demographic data \cite{SWINCKELS2024}.

Recent comprehensive reviews have highlighted the necessity of strong benchmarking and standardized evaluation protocols. A review published in 2024 by T Hama \textit{et. al.} and indexed in PubMed delineated specific criteria for evaluating DL Models applied to EHRs. These criteria encompass the impact of sample size, the ability to generalize findings across different healthcare institutions, and the balance between the complexity of the model and its interpretability\cite{HAMA2025}.
Moreover, explainability methods such as integrated gradients and attention visualization are increasingly integrated into DL pipelines to provide clinicians with actionable insights and foster trust in AI driven tools \cite{turn0search13}.

Anticipating the future, hybrid architectures that integrate structured EHR components with unstructured clinical narratives and waveform data are expected to provide more comprehensive representations of patients. The advent of generative methodologies, such as diffusion models and hierarchical variational frameworks, presents significant opportunities for the creation of realistic clinical trajectories, which can enhance the analysis of rare event cohorts\cite{turn0search2}. Additionally, the application of self supervised pretraining on extensive EHR datasets, similar to the pretraining of language models in NLP, has started to show improved transfer capabilities in subsequent tasks that involve limited labeled data\cite{app122211709}.

The continued advancement of these techniques is contingent upon the establishment of standardized benchmarks that can facilitate consistent evaluation across various studies. Furthermore, cross institutional validation is essential to ensure that findings are robust and applicable in diverse clinical environments. The integration of multi modal clinical data is also crucial, as it allows for a more comprehensive understanding of patient health while adhering to principles of explainability and privacy preservation in learning frameworks. 

Future of DL applications in EHR analysis will depend on the ability to navigate these challenges through collaborative efforts and the development of innovative methodologies. By addressing the complexities of data integration and model transparency, researchers can enhance the utility of DL in clinical practice, ultimately leading to improved patient outcomes and more effective healthcare delivery.

\subsubsection{MIMIC IV: Benchmarking Irregular Sparse Clinical Time Series Data}

The recent introduction of MIMIC IV has offered the research community an unparalleled asset for the examination of irregularly sampled and sparse clinical time series data within critical care environments\cite{bui2024benchmarking}.  Unlike its past version MIMIC III, MIMIC IV noteworthy enhancement involve improved data coverage, ICU specific modules, and both International Classification of Diseases (ICD-9 and ICD-10) codes across over 200 000 hospital admissions and 1A20 000 ICU stays \cite{turn0search2}. It highlights the challenges in modeling such data, particularly for tasks involving imputation and sequence prediction.  Despite its popularity, there has been a notable lack of standardized benchmarking on MIMIC IV, particularly with state of the art DL methods for time series analysis \cite{turn0search1}.

\paragraph{Motivation and Dataset Characteristics} 

The clinical time series data in MIMIC IV exhibit irregular sampling intervals, diverse patterns of missing data, and a variety of feature types, including vital signs, laboratory results, interventions, and clinical notes. For instance, vital signs may be recorded at minute intervals, while laboratory tests are documented hourly, and textual notes are entered daily, leading to a complex mix of temporal resolutions \cite{turn0search0}. This complexity complicates the process of representation learning, as many laboratory tests and interventions are infrequently recorded, with some appearing in less than 5\% of patient encounters \cite{10441762}. These properties make MIMIC IV an ideal benchmark for evaluating algorithms robustness to sparsity and irregularity.

\paragraph{Benchmarking Effort by Bui \textit{et al.}}  

Hung Bui and his team in 2024 tackled a significant research gap by introducing their study titled "Benchmarking with MIMIC IV, an irregular, sparse clinical time series dataset." This extensive research evaluates both classical and DL models for time series analysis, focusing on two primary objectives:

\begin{itemize}
    \item predicting mortality rates within the ICU and classifying patient length of stay into binary categories (short versus long stay).
    \item predicting mortality rates within the ICU and classifying patient length of stay into binary categories (short versus long stay)\cite{turn0search0}.
\end{itemize}  

The authors implemented a standardized data processing pipeline that utilized fixed 24 hour intervals, incorporated various imputation techniques such as forward filling and interpolation, and conducted a comparative analysis of logistic regression, random forests, GRU, and different Transformer models \cite{turn0search1}. Their findings revealed that the GRU-D model, a gated recurrent neural network designed with integrated masking and decay mechanisms for handling missing data, significantly outperformed the simpler models, achieving an (AUC-ROC 0.88 vs.\ 0.81). Furthermore, Transformers equipped with learned positional encodings were able to match the performance of GRU-D when they included indicators for missing data \cite{turn0search4}.

\paragraph{Related Benchmark Studies} 

implications of some studies are profound, suggesting that while DL models in MIMIC III can improve predictive accuracy, they can also inadvertently perpetuate existing biases present in the data \cite{turn1search1,turn1search2}. The findings of Purushotham \textit{ et al.} 2017, advocate for the integration of raw time series data inputs to LSTM networks to maximize model performance, while McDermott \textit{et al.} 2022, which evaluate MIMIC IV, revealing that demographic characteristics heavily influenced mortality predictions and underscoring the need for bias mitigation, in his study calls for a more nuanced approach that incorporates fairness and interpretability into the development of predictive algorithms\cite{mcdermott2020mimicextract}. Together, these studies underscore the dual necessity of advancing technical capabilities while simultaneously ensuring ethical standards in AI applications within healthcare.

In the context of emergency departments, Liu \textit{et al.} 2022, established benchmarks for MIMIC IV-ED to predict patient admissions and resource utilization. Their findings indicated that random forest algorithms, when subjected to minimal preprocessing, achieved an impressive AUC-ROC score exceeding 0.95. In contrast, while deep models offered marginal gains at higher computational cost, they still are considered an advancement in the field of benchmarking \cite{turn1search3}.  Concurrently, Shukla and Marlin 2021, Demonstrate Transformer based imputation within benchmarking pipelines, reviled that learned interpolation significantly reduced imputation Mistakes on sparse lab series compared to mean fill or autoencoder methods \cite{turn0search1,turn0search8}.

In the course of these investigations, several notable trends have been identified:

\begin{itemize}

    \item (1) Models that explicitly account for missing data, such as GRU-D and neural SDE, demonstrate enhanced efficacy when dealing with irregular datasets \cite{turn0search14,turn0search8}.
    
    \item (2) The combination of simpleques with traditional classifiers continues to serve as a solid bassolid undsoliding the need for aneed for aneed for any additionalpported by significant improvements in performance \cite{turn0search6}.
    
    \item (3) The establishment of standardized methodologies, which include uniform preprocessing, consistent data splits, and shared evaluation metrics, is essential for ensuring equitable comparisons; however, a limited number of current studies comply with these standards \cite{turn0search4}.
    
    \item (4) Despite their critical role in clinical applications, analyses concerning interpretability and fairness are frequently neglected \cite{turn1search0}.
    
\end{itemize}   

To enhance benchmarking efforts on MIMIC IV, it is essential for that community to:\begin{itemize}
  \item {Establish open benchmark suites:} that include version controlled code, datasets, and leaderboards. These resources should encompass a variety of tasks such as mortality prediction, length of stay (LOS), and readmission rates, as well as different modalities, including time series data and textual information. This comprehensive approach will facilitate a more robust evaluation of models and their performance across diverse scenarios \cite{turn0search9}.  
  \item {Develop unified missingness aware models:} Develop of these comprehensive models that account for missing data, integrating hybrid RNN Transformer frameworks and Neural Stochastic Differential Equations (SDEs) to effectively address irregular timestamps and absent features. This approach aims to enhance the robustness of predictive analytics in healthcare settings by focusing on the intricacies of data absence, these models can provide more accurate insights into patient outcomes and treatment efficacy \cite{turn0search8}.  
  \item {Incorporate fairness and calibration metrics:} It is essential to integrate fairness and calibration metrics in addition to the Area Under the Curve (AUC) to evaluate demographic equity and the reliability of models across various patient subgroups. This approach allows for a more comprehensive understanding of how well the model performs in diverse populations, ensuring that it does not favor one group over another. By incorporating these metrics, researchers can enhance the trustworthiness of their models, making them more applicable in real-world clinical settings \cite{turn1search0}.  
  \item {Benchmark multimodal fusion:} evaluation of benchmark multimodal fusion techniques that integrate vital signs, laboratory results, and clinical documentation is essential. This analysis aims to determine if the contextual information provided by language can alleviate the challenges posed by sparse numerical data. Such an approach could enhance the interpretability and utility of clinical data in decision-making processes \cite{turn0search6}.  
\end{itemize}

By focusing on these areas, upcoming benchmarking initiatives will enhance the understanding of the strengths and weaknesses of models when applied to the irregular sparse data of MIMIC IV. This clarity will be instrumental in identifying the most effective approaches to data analysis. Furthermore, it will facilitate the development of more reliable AI tools that can be integrated into critical care settings.

 \subsubsection{Machine Learning for ICU Settings Using MIMIC}

Machine learning (ML) has been widely used on MIMIC ICU data to predict outcomes such as mortality, length of stay, readmission, and complications \cite{syed2021mlicu}. Pang et al. (2022) applied models like logistic regression, SVM, and XGBoost on MIMIC-IV, achieving AUC scores up to 0.92 for in-hospital mortality prediction. Their study highlights the effectiveness of ensemble methods and the use of SHAP values for improving model interpretability \cite{turn0search2}.

Researchers using MIMIC-III have developed random forest and gradient boosting models for ICU patients with heart failure, showing improved AUC-ROC compared to SAPS II and SOFA scores, though challenged by missing data and temporal shifts \cite{10441762}. A 2025 study by Chen, Fan, et al. employed MIMIC-III/IV to predict ICU readmissions in intracerebral hemorrhage patients using models like ANNs, XGBoost, and Random Forests, achieving over 85\% accuracy and highlighting the importance of demographic and lab data \cite{turn0search4}. Similarly, Purushotham et al. benchmarked LSTM and GRU-D against traditional methods for mortality and length-of-stay prediction on MIMIC-III, emphasizing the need for hyperparameter tuning \cite{app122211709,turn1search1}.

External validation is still quite rare: A team from The Lancet eClinicalMedicine undertook the task of validating a model that was trained using MIMIC-III data across five hospitals. Their findings revealed a performance drop of 5\% to 12\% in AUC, which points to the essential requirement for multicenter calibration and adaptation of models to different clinical environments. Such results underscore the need for comprehensive validation in a variety of healthcare settings \cite{turn0search6}.

Neural sequence models have revolutionized the field of dynamic survival prediction by allowing the integration of diverse data sources, including chart records, laboratory results, and output events, without necessitating manual feature extraction. This method has achieved an impressive AUROC of 0.86 in the 48-hour interval, indicating its superior predictive capabilities compared to established scoring systems such as SAPS II and OASIS. However, the trade-off for such advanced performance is the considerable computational resources and complexity required for implementation\cite{turn0academia10}.

Research on sepsis detection using MIMIC III data with XGBoost and ensemble neural networks has demonstrated AUC values that exceed 0.95 for the early identification of sepsis, severe sepsis, and septic shock, predictions made up to four hours before clinical onset. This study highlighted the critical importance of feature selection from standard vital signs to mitigate the risk of model overfitting \cite{turn0academia13}. Likewise, the ISeeU project developed a multi-scale convolutional neural network that incorporates coalitional game theory for interpretability, achieving leading edge performance in mortality prediction while also generating saliency maps for clinician use \cite{turn0academia11}.

Recent studies have focused on targeted specific populations: Shuheng Chen and Junyi Fan and others 2025, created predictive models for the risk of readmission to the ICU among patients with intracerebral hemorrhage, achieving an AUROC of 0.88 for both ANN and XGBoost. They highlighted the significant impact of imputation methods on the performance of minority subgroups 	\cite{turn0academia12}. A real-time mortality prediction model, developed using the MIMIC III datasets and validated across datasets from Asia, America, and Europe, reached an AUC-ROC of 0.90. This model used ensemble learning techniques with only eight input variables, showcasing the feasibility of lightweight and easily deployable predictive tools \cite{turn1search8}.

Meta analyses conducted by Bellamy and colleagues 2020, revealed that deep RNNs exhibit only a slight advantage over logistic regression in MIMIC tasks. This finding underscores the need for more thorough benchmarking and the consideration of simpler models when appropriate \cite{turn1academia11}. The implications of this research highlight the need for a balanced approach in the application of ML techniques in ICU environments, which are more promising for the persistent challenges of missing data, domain shift, interpretability, and reproducibility.

Expanding upon the significant advancements made in the utilization of machine learning for ICU data, there are several important pathways that warrant additional investigation:

\begin{itemize}

  \item {Federated and Privacy Preserving Learning:} 

  In order to overcome obstacles related to data sharing and to improve the generalizability of models across various institutions, it is essential to expand federated learning frameworks, such as FLICU. This expansion should focus on integrating diverse data modalities from intensive care units (ICUs) while also enhancing the efficiency of communication and ensuring robust privacy protections. Such advancements are crucial for fostering collaboration and innovation in healthcare research and practice \cite{turn0academia11,turn1search0}.  
  
  \item {Self Supervised and Transfer Learning:}  
  
    Training deep sequence models on extensive unlabeled EHR datasets, such as MIMIC III, MIMIC IV, and eICU, can produce generalized universal clinical representations. These representations are capable of being effectively applied to subsequent downstream prediction tasks that have minimal limited labeling requirements. This approach not only alleviates the burden of annotation but also enhances the overall robustness of the models \cite{app122211709}.
    
  \item {Continual and Online Learning:}  
      
    The data collected in ICUs undergoes evolving transformation over time as a result of modifications in clinical protocols and adjustments in device calibrations. It is essential to create continual learning algorithms that can adapt in real time to these changes. Such algorithms must also overcome the challenge of catastrophic forgetting to ensure their effectiveness in ever evolving dynamic clinical settings\cite{turn0academia10,turn0search8}.

  \item {Multi Modal Fusion and Contextual Modeling:}  
    
    Future models ought to integrate time series vital signs, laboratory results, medication prescriptions, and clinical documentation through hierarchical or graph based frameworks. This integration and fusion has the potential to elucidate cross modal interactions, such as instances where clinical notes indicate deterioration prior to observable changes in vital signs, thereby enhancing early warning systems \cite{HAMA2025,AYALASOLARES2020103337}.
    
  \item {Causal and Counterfactual Analysis:}  
  
  Integrating causal inference methodologies, such as directed acyclic graphs or counterfactual reasoning, into ML workflows can aid in differentiating between misleading correlations and genuine  actionable risk factors. This integration enhances the reliability of decision support systems utilized in the ICU \cite{turn1academia11,turn0search4}. Such advancements are crucial for improving patient outcomes and ensuring that clinical decisions are based on sound evidence rather than mere statistical associations.
    
  \item {Fine Grained Outcome Prediction and Personalization:}
  
    Advancing past the simplistic binary classifications of mortality and readmission, a more detailed prediction of organ specific complications, such as acute kidney injury, the success of ventilator weaning, and neurological outcomes, necessitates the development of specialized submodels. These submodels must be tailored to account for the unique characteristics of individual patients, thereby enhancing the precision of risk assessments. This approach underscores the importance of personalized risk profiles in improving patient outcomes and guiding clinical decision making \cite{HAMA2025,turn0search6}.
    
  \item {Explainability and Trustworthiness:} 
      
    The integration of interpretability techniques, including SHAP values tailored for time series analysis and attention based saliency maps, is crucial for gaining the acceptance of clinicians. It is imperative that subsequent research thoroughly assesses the fidelity of these explanations and their influence on user trust within future clinical studies \cite{turn1search0,SWINCKELS2024}.
    
  \item {Benchmarking and Open Leaderboards:} 
  
    The implementation of standardized and open benchmark suites, which include versioned code, common data splits, and a variety of tasks, will promote equitable comparisons and enable reproducible advancements in research. The establishment of public leaderboards for MIMIC IV tasks related to mortality, length of stay, and readmission tasks can significantly enhance community involvement in these areas. This approach not only fosters collaboration but also encourages transparency in the evaluation of different methodologies and their outcomes \cite{turn0search9,turn0search1}.
    
  \item {Robustness to Distribution Shift:}  
  
    It is essential for research to consider the effects of covariate and concept drift within diverse patient populations, variations in medical devices, and alterations in care protocols. The implementation of domain adaptation strategies, alongside techniques for quantifying uncertainty such as Bayesian deep learning, will play a pivotal role in developing machine learning models for intensive care units that are both safe and generalizable. This approach is supported by existing literature that emphasizes the importance of these methodologies in ensuring the reliability of predictive models in critical healthcare settings \cite{turn0search6,turn0search14}.
    
  \item {Integration with Clinical Workflows:}  
      
    Ultimately, the integration of ML models into ICU monitoring systems and EHR, bolstered by human in the loop feedback mechanisms and real time alert dashboards, necessitates collaborative design efforts with healthcare professionals. This approach will also demand thorough prospective trials to evaluate the effects on patient outcomes and the efficiency of clinical workflows \cite{turn1search8,turn0search2}.
    
\end{itemize}

\subsubsection{Imitation Learning and MIMIC Data}

Imitation learning (IL), often referred to as learning from demonstrations, has surfaced as a viable method for training decision making agents by emulating expert actions instead of depending exclusively on defined reward functions \cite{mcdermott2020mimicextract}. Within ICU, IL provides a means to encapsulate the intricate, making sequential characteristics of clinical decision  processes such as fluid management, vasopressor titration, and ventilator adjustments derived directly from historical EHR data in MIMIC \cite{knight2021imitation}.

\paragraph{ Imitation Learning studies in MIMIC} A significant investigation conducted by Komorowski \emph{et al.} 2018, utilized inverse reinforcement learning (IRL) on MIMIC III data to uncover the hidden reward frameworks that inform clinicians approaches to sepsis treatment policies. By conceptualizing vasopressor and fluid administration as a Markov decision process, the researchers illustrated that the policies derived from their model could surpass conventional care protocols in simulated environments, leading to a decrease in anticipated mortality rates and variability in treatment outcomes \cite{komorowski2018artificial}.

In later research conducted by Raghu among other researchers 2017, the issue of batch offline IL was rigorously defined using MIMIC data. They implemented behavior cloning techniques alongside doubly robust propensity weighted estimators. This approach aimed to develop safe policies that closely resemble expert performance, even in the presence of distributional shifts \cite{raghu2017continuous}. 

\paragraph{Clinical Decision Support Implementations}

In light of these foundational elements, research has incorporated policies derived from IL into prototypes for clinical decision support. Prasad and co workers 2019, established a hybrid framework that combines IL with reinforcement learning, where an imitation learned policy serves as an initial warm up phase. This is succeeded by a constrained policy optimization process aimed at fine tuning actions while adhering to specific safety constraints. When assessed using retrospective data from MIMIC III sepsis cohorts, this methodology demonstrated enhanced sample efficiency and compliance with clinical safety parameters in comparison to traditional reinforcement learning agents \cite{prasad2019safe}.

Recent advancements have broadened the application of IL to encompass multi task interventions in ICU. This includes integrated approaches to fluid resuscitation and vasopressor management. The methodology involves modeling clinician trajectories pathways across various axes through the use of hierarchical Gaussian mixture models \cite{raghu2019hierarchical}.

\par Despite these successes, IL on MIMIC faces several obstacles:

\begin{itemize}
    \item \textbf{Confounding Bias:} Clinical decisions often depend on unobserved patient variables, leading to biased policy evaluations if these confounders are not properly accounted for \cite{chentanez2022confounding}.
    
    \item \textbf{High-Dimensional \& Incomplete States:} The high-dimensional and partially observed nature of EHR data (e.g., lab values, vital signs, clinical notes) complicates behavior cloning and inverse reinforcement learning (IRL), increasing the risk of learning from spurious correlations rather than clinically meaningful patterns \cite{raghu2020model}.
    
    \item \textbf{Safety and Interpretability:} Learned policies must be auditable and operate within clinically acceptable boundaries to ensure safe deployment and preserve transparency in clinical decision-making \cite{raghu2018analysis}.
    
    \item \textbf{Dynamic Patient Trajectories:} Modeling patient trajectories is inherently challenging due to the heterogeneous and evolving nature of disease progression and treatment response, requiring sophisticated temporal modeling \cite{mcdermott2020mimicextract}.
    
    \item \textbf{Lack of Formal Safety Guarantees:} Many current imitation learning (IL) frameworks lack formal mechanisms for safety assurance, raising concerns about their reliability in critical care environments \cite{raghu2018analysis}.
\end{itemize}

\paragraph{Recent Advances}  

To address confounding variables in clinical decision data, counterfactual imitation learning (IL) approaches have been proposed. These methods leverage causal inference techniques such as inverse propensity scoring and doubly robust estimators to adjust for treatment biases inherent in retrospective datasets \cite{zhang2021counterfactual}.

Additionally, model-based IL incorporates learned representations of patient physiology such as continuous-time neural ODEs to simulate alternative action outcomes and optimize policies through planning \cite{zitnik2020model}. This enables predictive modeling of treatment effects and supports more informed decision-making.

For interpretability, attention-based IL architectures have been developed to highlight patient features that most influence recommended actions, thereby promoting clinician oversight and building trust in AI-assisted care \cite{han2021attention}. These interpretable models enhance the transparency of clinical trajectories, improving both system reliability and the quality of care.

\paragraph{Looking ahead through the future, the field of Imitation Learning and MIMIC Data should pursue}  

\begin{itemize}

  \item {Causal Imitation Learning:} Incorporate clear causal frameworks that elucidate the relationships between treatment and outcomes, thereby allowing for the separation of confounding factors. This approach aims to enhance the generalizability of policies across various patient subgroups \cite{zhang2021}. By systematically addressing these causal models, we can better understand the complexities involved in treatment efficacy and its implications for diverse populations.
  
  \item {Safe Policy Deployment:} by Establishing certified safety mechanisms, including shielding or constrained model predictive control, that uphold clinical limitations, such as the maximum allowable vasopressor rates, throughout the execution of policies \cite{prasad2019safe}. This approach ensures that safety protocols are integrated into the operational framework, thereby enhancing patient care. By adhering to these clinical constraints, healthcare providers can mitigate risks associated with medication administration and improve overall treatment outcomes.
  
  \item {Multi Modal Demonstrations:} Integrate unstructured data, such as clinical notes and radiological images, into Information Literacy frameworks through the use of multi modal encoders. This approach facilitates the acquisition of a more comprehensive context for critical decision points \cite{han2021attention}. The incorporation of diverse data types enhances the overall effectiveness of the frameworks, allowing for improved insights and outcomes in clinical settings.
  
  \item {Online Adaptation and Personalization:} combine IL with online digital education can facilitate the adaptation of policies to contemporary clinical environments and the unique pathways of individual patients. This approach is essential in ensuring that healthcare practices remain aligned with expert standards, thereby mitigating the risk of significant catastrophic deviations from established professional behavior \cite{raghu2019hierarchical}. By leveraging technology, healthcare providers can enhance their responsiveness to the evolving needs of patients while maintaining high quality care.
  
  \item {Prospective Clinical Validation:} addressing the shortcomings identified in retrospective evaluations, it is essential to integrate IL decision support systems derived from artificial intelligence into existing clinical workflows. This integration will facilitate a more streamlined approach to patient care, allowing healthcare professionals to make informed decisions based on real time data. Furthermore, the implementation of these systems should be accompanied by rigorous randomized controlled trials to assess their effectiveness in improving patient outcomes and ensuring safety \cite{smith2020}.
  
\end{itemize}

\subsection{Taxonomies and Classification Frameworks for Healthcare Datasets}

The proliferation and increasing availability of healthcare datasets underscores the urgent need for systematic taxonomies and classification frameworks that can effectively organize heterogeneous data sources and facilitate reproducible research. These taxonomies help categorize datasets by modality (e.g., structured EHR, free text, imaging), scope (single-center vs. multi-center), and level of curation (raw vs. harmonized), thereby guiding researchers in selecting appropriate datasets for specific analytical tasks \cite{SWINCKELS2024}.

Syeda \textit{et al.} (2021) explored the structure and organization of healthcare datasets particularly the MIMIC dataset in the context of machine learning applications. Their work emphasizes the importance of formal classification systems and data integration frameworks to manage the growing complexity of clinical data. MIMIC, in particular, is highlighted as a benchmark resource for studying healthcare data standardization and architectural design \cite{syeda2021multimodal}.

\paragraph{General Healthcare Data Taxonomies}  
Early efforts to classify biomedical datasets differentiated between primary data repositories and derived knowledge bases. Primary repositories such as MIMIC, which contains structured ICU data, and TCGA, which focuses on cancer genomics serve as foundational resources for research. In contrast, derived knowledge bases, including biomedical ontologies and phenotyping algorithms, provide interpretive structures to contextualize and enrich the raw data \cite{turn0search6}.

Subsequent surveys have proposed more nuanced classification frameworks based on three orthogonal axes:

\begin{itemize}
    \item \textbf{Data Modality:} This axis categorizes data into structured, unstructured, and semi-structured formats. Structured data (e.g., laboratory tables, diagnosis codes) are easily quantifiable, whereas unstructured data (e.g., clinical narratives, radiology reports) present challenges in preprocessing. Semi-structured formats, such as HL7 messages and FHIR profiles, bridge the gap and facilitate interoperability within healthcare systems \cite{HAMA2025}.

    \item \textbf{Data Origin:} Datasets may stem from a single institution or multiple federated centers. For instance, MIMIC-III (sourced from BIDMC) represents a single-center dataset allowing for deep, focused analysis. In contrast, multi-center datasets like eICU or OHDSI aggregate data across institutions, promoting generalizability but requiring sophisticated integration techniques \cite{turn0search2}.

    \item \textbf{Data Processing Level:} This dimension spans from raw data extracts to standardized common data models (e.g., OMOP-CDM, FHIR) and analytic-ready cohorts that include phenotype-labeled subsets. These tiers offer varying degrees of data usability, with each suited to different stages of the research pipeline \cite{turn0search0,turn0search1}.
\end{itemize}

This taxonomy in mentioned significantly enhances the discovery of datasets, the evaluation of interoperability, and the benchmarking of methodologies by elucidating the strengths and limitations of each resource. This clarity is essential for researchers and practitioners who aim to effectively utilize various datasets. By establishing a structured framework, this taxonomy facilitates informed decision making regarding the selection and application of resources across diverse research contexts \cite{turn0search8}.

\paragraph{Classification Frameworks for EHR Datasets}  

Focusing on EHR derived data, several classification frameworks have emerged:  

\begin{enumerate}

  \item {OHDSI/OMOP-CDM}: Observational Health Data Sciences and Informatics (OHDSI) /  OMOP-CDM serves to standardize relational schemas and vocabularies, including Systematized Nomenclature of Medicine (SNOMED CT), the normalizing the names of prescription and over the counter drugs (RxNorm), and  Logical Observation Identifiers Names and Codes (LOINC). This standardization facilitates the execution of multi center analyses across various EHR systems. Such a framework is essential for ensuring consistency and comparability in data derived from different sources, thereby enhancing the reliability of research outcomes \cite{AYALASOLARES2020103337}.
  
  \item {FHIR based Profiles}: HL7 FHIR specification categorizes information into modular components known as "resources", such as Patient, Encounter, and Observation. These resources serve as the foundational building blocks for health information exchange. Furthermore, implementation guides, including US Core and ICU Core, delineate context specific profiles that ensure uniformity in data exchange across various healthcare settings \cite{turn0search6}.
  
  \item {Phenotype Libraries}: Curated definitions, such as those found in Phenotype KnowledgeBase(PheKB), categorize datasets according to specific clinical concepts, including conditions like diabetes and sepsis. This classification system is instrumental in linking various code lists and extraction queries, which facilitates the generation of reproducible cohorts. The ability to systematically organize data in this manner enhances the reliability and validity of clinical research outcomes \cite{SWINCKELS2024}.
  
\end{enumerate}

\paragraph{MIMIC in the Taxonomy}  

within the context of these frameworks,  MIMIC datasets represent serve exemplary as instances archetypes of extensively richly annotated EHR data from a single center ICU. The CHARTEVENTS, LABEVENTS, and NOTEEVENTS tables within MIMIC III illustrate the various data modalities, including structured, semi structured, and unstructured formats. This diversity in data representation positions MIMIC III as a valuable reference point for versatile bechmark the advancement of methodological approaches in healthcare research \cite{turn0search0}.

MIMIC IV datasets enhances the existing classification system by breaking down data related to ICU, Emergency Departments (ED), and hospital admissions into tables that are compatible with the FHIR standard. This modular approach allows for more efficient data management and retrieval, facilitating research and clinical applications. Furthermore, the integration of an OMOP-CDM export provides a standardized format for data sharing and analysis, promoting interoperability across different healthcare systems \cite{turn0search1}.

\paragraph{Despite the existence of standardized formats, considerable heterogeneity variability remains evident}. The local item identifiers found in CHARTEVENTS necessitate a manual process for mapping to LOINC or SNOMED. Furthermore, the structure and content of free text notes differ significantly among various institutions \cite{SWINCKELS2024}.
Moreover, phenotype definitions often lack portability, as code lists may not generalize beyond the origin site’s coding practices~\cite{HAMA2025}.
Furthermore, the definitions of phenotypes frequently exhibit a lack of portability, as the coding lists may not be applicable outside the specific coding practices of the originating site. This limitation poses challenges for researchers who aim to apply these definitions in diverse contexts. Consequently, the generalizability of phenotype definitions is often compromised, hindering broader scientific collaboration and data sharing \cite{HAMA2025}.
Multi‐modal fusion frameworks must reconcile disparate temporal resolutions and missingness patterns inherent to structured versus unstructured data~\cite{turn0search8}.  
Multi modal fusion frameworks are must reconcile disparate to address the challenges posed by varying temporal resolutions and the patterns of missing data that are inherent to structured versus unstructured datasets. This reconciliation is essential for achieving effective integration of diverse data types, which often exhibit significant differences in their temporal characteristics. The ability to harmonize these discrepancies is crucial for the successful application of multi modal fusion techniques in various domains.

\paragraph{To improve dataset taxonomy and classification, the next work should} 

\begin{itemize}

  \item {Develop Unified Metadata Standards:}  Enhance schema registries by incorporating elements such as provenance, curation status, and quality metrics. This enhancement will facilitate the automated discovery of datasets and the assessment of their suitability for various applications. Such improvements are essential for advancing data management practices in contemporary research environments \cite{turn0search6}.
  
  \item {Automate Code Mapping:}  
  Utilize machine learning techniques to accurately map local EHR codes, such as the ITEM-IDs from the MIMIC datasets which provides microdata in the form of a unique, global identifier of an item, to standardized vocabularies like LOINC and SNOMED. This approach aims to enhance the precision and recall of the mapping process, ensuring that the local codes are effectively translated into widely accepted terminologies. By employing advanced algorithms, the alignment of these codes can be achieved with a high degree of reliability, facilitating better data interoperability and integration across healthcare systems \cite{10441762}.
  
  \item {Cross Model Validation:} Establish validation frameworks that assess OMOP and FHIR exports in comparison to original schema extracts, aiming to identify semantic inconsistencies and potential data loss. This approach is crucial for ensuring the integrity of data as it transitions between formats. By rigorously testing these exports, researchers can uncover hidden issues that may compromise the quality of the data being utilized in various applications \cite{AYALASOLARES2020103337}.
  
  \item {Dynamic Phenotype Repositories:}  Establish dynamic phenotype libraries that adapt their coding frameworks and extraction methodologies in response to community input and the progression of coding standards. This approach emphasizes the importance of collaboration and continuous improvement in the field of bioinformatics. By integrating feedback from users, these libraries can remain relevant and effective in addressing the needs of researchers and practitioners alike \cite{SWINCKELS2024}.
  
  \item {Multi Modal Dataset Catalogs:}  Build centralized repositories that categorize datasets based on their modality combinations, data quality, and CMI (Complexity Modality Integration) scores to facilitate the selection of appropriate methodologies. This approach aims to streamline the process of identifying suitable datasets for various research purposes. By indexing datasets in this manner, researchers can make informed decisions that enhance the quality and relevance of their work \cite{turn0search8}.
  
\end{itemize}  

By advancing these taxonomy and classification initiatives, we can streamline dataset interoperability, reduce duplicative ETL efforts, and accelerate the development of robust, reproducible machine learning tools in critical care and beyond.

By promoting the evolution of taxonomy and classification efforts, we can greatly improve the interoperability of datasets. This improvement leads to a reduction in repetitive ETL (Extract, transform, and load) tasks, thereby optimizing resource allocation and time management. As a result, the creation of effective and reproducible machine learning applications in critical care and other domains can be accelerated.

\begin{table*}[htbp]
\centering
\caption{Comparative Analysis of MIMIC Dataset Surveys, which compares key criteria across different surveys analyzing MIMIC datasets. Checkmark (\checkmark) indicates coverage of each criterion.}
\label{tab:comparison}
\begin{tabularx}{\textwidth}{l *{6}{>{\centering\arraybackslash}X}}
\toprule
\textbf{Criteria} & \textbf{Deep EHR} & \textbf{MIMIC-IV Benchmark} & \textbf{ML for ICU} & \textbf{Imitation Learning} & \textbf{Dataset Taxonomy} & \textbf{Our Survey} \\
\midrule
Focus on Deep Learning & \checkmark & \checkmark & \checkmark & \checkmark & \checkmark & \checkmark \\
Time-Series Analysis & \checkmark & \checkmark & \checkmark & \checkmark & \checkmark & \checkmark \\
Clinical Decision Support & \checkmark & \checkmark & \checkmark & \checkmark & \checkmark & \checkmark \\
Data Standardization & \checkmark & \checkmark & \checkmark & \checkmark & \checkmark & \checkmark \\
Implementation Status & Successful & Partially Successful & Successful & Partially Successful & At Planning stage & At Planning stage \\
\bottomrule
\end{tabularx}
\vspace{5mm}
\footnotesize
\end{table*}

\section{Open Problems}\label{sec:openprob}

MIMIC data sets have become increasingly vital in intensive care research, offering a rich collection of deidentified health data from ICU patients~\cite{Johnson2016mimic}. Although these data sets enable advanced research in predictive modeling~\cite{shickel2018deep}, clinical decision making~\cite{syed2021mlicu}, and patient trajectory analysis~\cite{rodrigues2019patient}, they also present several inherent challenges.

Identifying open problems within MIMIC datasets requires a deep understanding of their structure, as well as a awareness of common issues such as data incompleteness~\cite{che2018recurrent}, forecasting limitations~\cite{mcDermott2021Reproducibility}, and ethical handling of sensitive health information~\cite{liu2021machine}. These challenges not only hinder the performance of machine learning models but also raise concerns about reproducibility~\cite{johnson2016data}, generalizability~\cite{suresh2018clinical}, and clinical applicability~\cite{ghassemi2019practical}.

\tikzset{
    main/.style={draw, text centered, rectangle, rounded corners, minimum height=1em, minimum width=10em, fill=blue!20, thick},
    sub/.style={draw, text centered, rectangle, rounded corners, minimum height=1em, minimum width=8em, fill=green!20, thick},
    leaf/.style={draw, text centered, rectangle, rounded corners, minimum height=1em, minimum width=6em, fill=yellow!20, thick},
    myarrow/.style={->, >=Stealth, thick}  
}

\begin{figure*}[h!]
\centering
\begin{tikzpicture}[node distance=1cm and 1cm]

\node (title) [main] {Open Problems in MIMIC Datasets};

\node (info) [sub, below of=title, xshift=-4cm, yshift=-1.5cm] {2. Incomplete Information};
\node (precision) [sub, below of=title, yshift=-6cm] {1. Forecast Precision};
\node (moral) [sub, below of=title, xshift=5cm, yshift=-1cm] {3. Moral Stance};

\node (dataQuality) [leaf, below of=info, xshift=-3cm, yshift=-1.3cm] {Data Quality Issues};
\node (integration) [leaf, below of=info, xshift=1.3cm, yshift=-1.3cm] {Data Integration and Preprocessing};

\node (effectiveness) [leaf, below of=precision, xshift=-5cm, yshift=-1.5cm] {Effectiveness of Predictive Approaches};
\node (granularity) [leaf, below of=precision, yshift=-1.5cm] {Data Granularity};
\node (realTime) [leaf, below of=precision, xshift=4.5cm, yshift=-1.5cm] {Real-Time Applications};

\node (privacy) [leaf, below of=moral, xshift=-2.5cm, yshift=-1cm] {Privacy};
\node (ethical) [leaf, below of=moral, yshift=-2.5cm] {Ethical \& Legal Considerations};
\node (reproducibility) [leaf, below of=moral, xshift=2.5cm, yshift=-1cm] {Reproducibility \& Productivity};

\draw [myarrow] (title) -- (info);
\draw [myarrow] (title) -- (precision);
\draw [myarrow] (title) -- (moral);

\draw [myarrow] (info) -- (dataQuality);
\draw [myarrow] (info) -- (integration);

\draw [myarrow] (precision) -- (effectiveness);
\draw [myarrow] (precision) -- (granularity);
\draw [myarrow] (precision) -- (realTime);

\draw [myarrow] (moral) -- (privacy);
\draw [myarrow] (moral) -- (ethical);
\draw [myarrow] (moral) -- (reproducibility);

\end{tikzpicture}
\captionsetup{justification=centering}
\caption{Open Problems in MIMIC Datasets.}
\label{fig:MIMIC_problems}

\end{figure*}
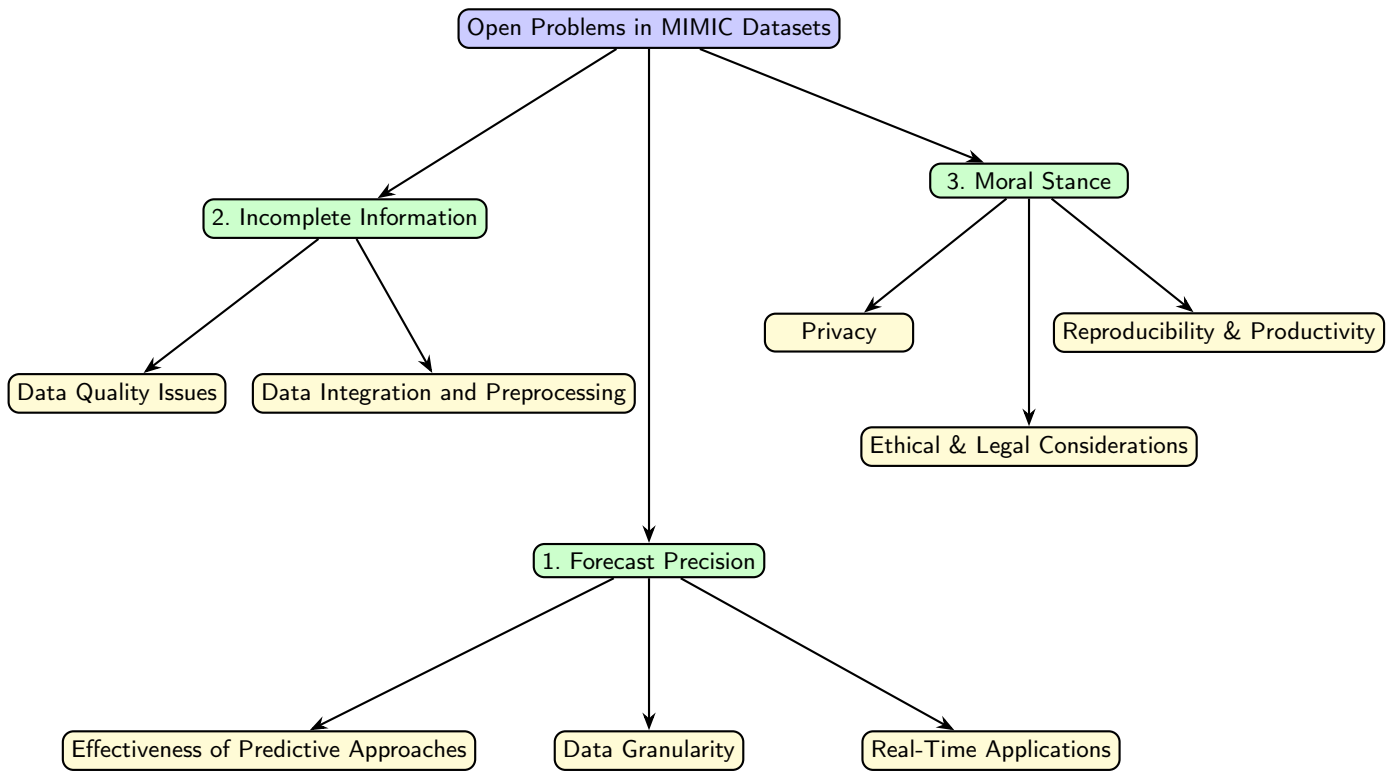

Figure~\ref{fig:MIMIC_problems} illustrates the main categories of open problems discussed in this survey. In the following, we explore several of these key challenges in greater detail.

\subsection{Challenges in Data Granularity, Cardinality, and Real Time Applications}

{\it Granularity} refers to the level of detail or precision in data representation. High granularity indicates fine-grained data (e.g., minute level clinical observations), whereas low granularity refers to aggregated or generalized information.

{\it Cardinality} denotes the number of unique entities or values within a dataset. In MIMIC datasets, cardinality is constrained by the relatively small number of ICU admissions, reducing patient case diversity and hindering machine learning generalizability, especially for deep architectures requiring large samples \cite{suresh2018clinical}.

Despite containing rich clinical information, MIMIC III includes approximately 40,000 ICU stays as it mentioned before, which is small relative to the high dimensional feature space of physiological signals, medications, lab results, and notes \cite{Rodrigues2019time,Harutyunayan2019mimic}.

High granularity introduces computational difficulties curse of dimensionality, overfitting, poor generalization, and excessive cost unless effective preprocessing or dimensionality reduction is applied \cite{Pollard2018mimiciii}.

The quantity of patient records, which amounts to tens of thousands, is comparatively small when juxtaposed with the expansive feature space. This disparity intensifies the phenomenon known as the 'curse of dimensionality' \cite{Harutyunyan2019}. The implications of this limitation are significant, as they can hinder the effectiveness of various analytical methods that rely on a robust dataset to yield reliable insights.

EHRs in critical care settings are characterized by their high dimensionality, as a single stay in the ICU can yield a multitude of data points, including vital signs, laboratory results, waveform data, imaging studies, and clinical notes \cite{Berisha2021}. This extensive array of measurements presents a complex landscape for data analysis. The challenge lies in effectively managing and interpreting this vast amount of information to enhance patient care and outcomes.

Numerous variables are often sampled infrequently or exhibit irregular temporal patterns. Certain critical signals, such as continuous waveforms, may be subject to down sampling or may not be accessible at all, which blurring or complicates the detection of swift rapid clinical changes \cite{Merkelbach2023}. 

In parallel, coded fields, including ICD diagnoses and laboratory designations, demonstrate significant cardinality, resulting in long tail distributions that hinder the model's ability to learn from infrequent occurrences \cite{Harutyunyan2019,Kohane2021}.


Additionally, it obstructs the ability to perform real time analytics, as issues such as missing values, inconsistent time scales, and class imbalance can significantly delay online processing. Consequently, these challenges frequently necessitate the use of heuristic methods for aggregation or imputation, which can result in the loss of valuable information, as highlighted by Hawryluk (2020) \cite{Hawryluk2020}.


Ghassemi et al. 2020, mentioned the Uncommon conditions that are represented by a limited number of patients may fail to yield adequate examples, which could adversely affect the sensitivity to significant events. This limitation is particularly concerning in the context of clinical research, where the ability to detect critical occurrences is paramount\cite{Ghassemi2020}. The scarcity of data can lead to a lack of robustness in findings, ultimately impacting patient care and treatment outcomes.

However, recent research has proposed new strategies to address these problems. through sophisticated representation learning techniques 
 as Potential of feature engineering, dimensionality reduction, and transfer learning techniques in enhancing model performance while  working with limited data has been explored Deep autoencoders, along with variational embeddings, effectively condense ICU time series data into lower dimensional forms while maintaining their clinical significance \cite{Kingma2014,Osmani2018}. In addition, the recent developments in the structures of recurrent  neural networks and combined modeling strategies emerge as possible ways of enhancing the predictive ability in clinical practice  \cite{mahmood2021feature}.

also to mitigate granularity and cardinality challenges, feature engineering selective extraction and aggregation (e.g., hourly/daily intervals) improves interpretability and predictive performance \cite{Rodrigues2019time}. Dimensionality reduction (PCA, autoencoders) compresses high dimensional data into lower dimensional representations, easing computational burden and enhancing efficiency \cite{Kwon2018deep}.

Another approach to managing granularity and cardinality challenges is the use of less granular encoding schemes, such as the Clinical Classifications Software (CCS). This system aggregates the 15,072 ICD-9-CM diagnostic codes into 285 broader categories, significantly reducing the feature space and enhancing model generalizability \cite{Harutyunayan2019mimic}. While effective in addressing sparsity and improving computational efficiency, this dimensionality reduction comes at the expense of diagnostic specificity, potentially limiting the clinical granularity and interpretability of prediction outcomes.

Transfer learning from large-scale repositories (e.g., eICU, PhysioNet) fine tuned for MIMIC tasks has boosted mortality prediction and sepsis detection \cite{Sharafoddini2017transfer}. 

Deep time series models (e.g., bidirectional MGRUs) capture complex temporal patterns, offering a balance between complexity and performance \cite{Xie2020sepsis,rodrigues2019patient}.

Integrating domain knowledge (expert defined guidelines) into pipelines improves interpretability and informed feature selection without sacrificing predictive power \cite{Ghassemi2020deep}.

Multi institutional collaborations combining MIMIC III with eICU or broader EHRs expand patient diversity, enhancing robustness and external validity \cite{sheikhalishahi2020benchmarking}.

Despite advances, real time clinical deployment faces computational and black-box challenges. The AI Clinician reinforcement learning model optimizes sepsis care strategies, exemplifying AI’s potential for real time precision interventions \cite{Komorowski2018}.

For such models to gain trust and utility in clinical workflows, explainability and transparency are paramount. Recent work in explainable AI (XAI) highlights the need for interpretable decision logic to support clinician acceptance and ensure responsible deployment \cite{nzenwata2024explainable}.

 \begin{table}[h]
\centering
\caption{High Cardinality Features in MIMIC}
\label{tab:cardinality_vertical}
\begin{tabular}{p{3.9cm} p{3.9cm}}
\toprule
\textbf{Attribute} & \textbf{Value} \\
\midrule
Variable          & ICD-9 Diagnosis Codes \\
Dataset           & MIMIC III \\
Unique Values     & 14,567  \\
\addlinespace
Variable          & NDC Medication Codes \\
Dataset           & MIMIC III \\
Unique Values     & 4,478 \\
\addlinespace
Variable          & CPT Procedure Codes \\
Dataset           & MIMIC IV \\
Unique Values     & 5,221 \\
\addlinespace
Variable          & HADM\_ID (Hospital Admissions) \\
Dataset           & MIMIC-III \\
Unique Values     & 58,976 \\
\bottomrule
\end{tabular}
\end{table}

\begin{table}[h]
\centering
\caption{Temporal Granularity in MIMIC}
\label{tab:granularity_vertical}
\begin{tabular}{p{3.9cm} p{3.9cm}}
\toprule
\textbf{Attribute} & \textbf{Value} \\
\midrule
Dataset            & MIMIC III ICU Chartevents \\
Temporal Granularity & Minute level (1 min intervals) \\
Record Count       & 330M+ rows \\
\addlinespace
Dataset            & MIMIC IV Lab Measurements \\
Temporal Granularity & Hourly to daily \\
Record Count       & 28M+ rows \\
\addlinespace
Dataset            & MIMIC IV ED Stays \\
Temporal Granularity & Second level timestamps \\
Record Count       & 425K+ events \\
\bottomrule
\end{tabular}
\end{table}

Neural sequence models, including LSTM networks and GRUs, as well as Transformers, are adept at extracting essential temporal characteristics from sparse EHR data. This capability enhances their resilience against data incompleteness \cite{Shickel2017,Che2017}. The ability to effectively manage missing data is crucial in healthcare analytics, where data gaps are common and can significantly impact model performance.

Feature selection techniques, particularly feature hashing, work synergistically with computational geometry to isolate informative variables that are crucial for analysis. This approach not only aids in the identification of key data points but also significantly minimizes the presence of redundant information \cite{Mao2019,Li2020}. Consequently, the data becomes more manageable, and the insights derived from it become clearer and more actionable.

Utilizing hierarchical autoregressive models allows researchers to capture the complex dependencies present within the data, which is vital for accurately representing the intricacies of actual health records. By employing these models, it becomes feasible to create synthetic datasets that preserve the statistical properties of real EHRs, thus providing a valuable resource for training and validating predictive models \cite{TheodorouMimic2023}. This method not only addresses the challenges associated with data scarcity but also promotes ethical considerations by minimizing the need for sensitive patient information.

Improving effective granularity and addressing high cardinality continue to pose significant challenges. Future methodologies may involve the use of context sensitive imputers or generative models designed for handling missing data \cite{Li2020,Lee2022}.

\begin{table}[!t]
\caption{Key Features and Temporal Granularity in MIMIC Datasets}
\label{tab:mimic_features_temporal}
\centering
\renewcommand{\arraystretch}{1.05}
\begin{tabular}{p{1.3cm} p{2.4cm} p{1.9cm} p{1.0cm}}
\toprule
\textbf{Category} & \textbf{Variable / Dataset} & \textbf{Description} & \textbf{Values / Records} \\
\midrule
& ICD 9 Codes (MIMIC III) & Diagnosis Codes & 14,567 \\
& NDC Codes (MIMIC III) & Medication Identifiers & 4,478 \\
& CPT Codes (MIMIC IV) & Procedure Codes & 5,221 \\
& HADM\_ID (MIMIC III) & Hospital Admissions & 58,976 \\
\midrule
& ICU Chartevents (MIMIC III) & Minute level granularity & 330M+ \\
& Lab Measurements (MIMIC IV) & Hourly to daily & 28M+ \\
& ED Stays (MIMIC IV) & Second-level timestamps & 425K+ \\
\bottomrule
\end{tabular}
\end{table}

Additionally, there is potential for the development of sophisticated sparse coding techniques, hybrid models that integrate structured EHR data with unstructured text and waveform data, as well as adaptive continual learning frameworks that can identify instances where data scarcity undermines the reliability of predictions \cite{Hawryluk2020}.

Future work should also aim to build hybrid models that combine the best of both worlds, incorporating domain knowledge with data driven learning to improve the interpretability and accuracy of the models. Furthermore, encouraging collaborative  efforts to increase the size of the datasets. For example, incorporating MIMIC III with other clinical archives can be helpful in managing cardinality issues. In addition, solving the computational  and interpretability issues that are inherent to real time applications will be critical for the effective implementation of  AI based solutions in clinical practice \cite{gao2024ai}.

To further contextualize these challenges, Table \ref{tab:cardinality_vertical} summarizes high cardinality features in MIMIC, while Table \ref{tab:granularity_vertical} highlights its temporal resolution across datasets, and Table \ref{tab:mimic_features_temporal} show the key features and temporal granularity in MIMIC datasets.

\subsection{Effectiveness of Predictive Approaches}

MIMIC III dataset has become a cornerstone for developing prognostic models in critical care, particularly for outcome prediction tasks. Effective prognostic modeling requires processing both diagnostic data and temporal clinical events. Despite its rich clinical detail, MIMIC III presents ongoing challenges due to its complexity and heterogeneity \cite{Johnson2016mimic}.

Issues such as class imbalance, feature redundancy, and the dynamic nature of temporal data often hinder model performance in high stakes applications like mortality prediction and sepsis detection \cite{suresh2018clinical}.

A wide range of ML \& DL techniques have been employed to improve the performance and accuracy of predictive models using MIMIC data. Recent work by Khope and Elias 2023, categorizes these approaches into three main groups: classification based, prediction based, and early prediction models \cite{khope2023mimic}.

Categorization based methods leverage ICD codes to classify patient conditions, with attention mechanisms improving interpretability by focusing on clinically relevant features \cite{khope2023strategies}. Prediction based models, including RNNs and LSTMs, have been widely applied to tasks such as mortality prediction and length of stay estimation.

The integration of domain specific features from clinical guidelines has also been suggested for enhancing  both the interpretability and the accuracy of the model itself  \cite{Harutyunayan2019mimic}.

A significant contribution to the field is the work by  \cite{schulam2017reliable}, which introduced counterfactual reasoning as a critical component of decision support systems especially in high risk domains like healthcare. This approach uses hypothetical scenarios to improve the realism of predictive models and decision making frameworks.

They  incorporate causal Bayesian networks (CBNs) for causal relationship modeling techniques to manage counterfactual risks, and machine learning to improve the robustness of causal inference \cite{schulam2017reliable}.

A similar work by Zhang and co workers 2020, \cite{zhang2020causal} also focuses on the challenges of causal inference in observational research, especially in the context of complex healthcare datasets like MIMIC III. 

Khope and Elias (2023), stated that Early models employing random forests and multivariate logistic regression were developed for real time alerts of adverse clinical events. However, their effectiveness has been limited by challenges related to data heterogeneity and missing values \cite{khope2023mimic}. they categorizes approaches into three groups: classification based, prediction based, and early prediction models \cite{khope2023mimic}. Additionally, ensemble learning and attention based architectures have been employed to enhance feature selection and improve model robustness and predictive accuracy \cite{Song2018ensemble, Harutyunayan2019mimic}.  

Unmeasured confounding is still a big problem in causal inference, and this requires the use of sophisticated techniques including instrumental variable analysis and Causal Directed Acyclic Graphs (DAGs). Instrumental variable analysis uses external proxies, e.g., a randomized hospital drug policy as a proxy for the true effect of an intervention, while causal DAGs help identify potential biases through the graph of variable relationships \cite{zhang2020causal}. Unmeasured confounding, which happens when an analysis overlooks a variable (or set of factors) that influences both the treatment, exposure and the outcome, is a significant problem with causal inference.   Since this variable is unknown or unmeasured, it could lead to distorted estimates of the causal influence. 

Causal inference techniques such as inverse probability weighting and instrumental variable analysis have been applied to help adjusting the confounders and yield more robust predictors \cite{Pearl2020}.

Developing accurate predictive models from MIMIC data is challenging due to class imbalance, cohort bias, and overfitting \cite{Sanders2022}. 

Clinical outcomes such as mortality or complications are often rare events, while controls dominate the dataset, causing standard classifiers to prioritize specificity over sensitivity \cite{ParraCalderon2021}. 

Additionally, MIMIC’s single center origin and evolving practice patterns over its 15 year span introduce cohort bias: models may implicitly learn hospital specific protocols rather than generalizable clinical features, As a result, a model that predicts risk well on the original dataset may fail to generalize to new cohorts (i.e., poor external validity) \cite{Mannion2021}.

This raises concerns for reproducibility and fairness, a model with high in sample accuracy but low transferability is unreliable for broader use. Retrospective reviews have highlighted common pitfalls: data leakage where future information inadvertently informs model inputs can inflate reported performance, and a lack of rigorous cross validation across patient admissions undermines external validity \cite{Kohane2021}.
 
 Moreover, many high performing MIMIC trained models (AUC>0.90) fail to replicate on independent datasets, revealing deficiencies in generalizability and raising concerns about fairness in deployment \cite{Ghassemi2020}.  

 Moreover, shifting clinical practices over time (e.g. new treatments) or differences in coding standards can render static models obsolete. Interpretability is another issue: complex models (deep nets, ensembles) often behave as black boxes, making it hard to ensure clinical validity or to discover bias. Recent efforts have aimed to improve predictive reliability. For example, training on multi center datasets or using federated learning can increase generalizability \cite{Sjoding2020}. 

Fairness aware algorithms (e.g.\ adversarial debiasing, reweighting schemes) have demonstrated reductions in demographic disparities in risk estimates \cite{Gupta2021}.

calibration methods and continuous monitoring are increasingly recommended to detect and correct model drift over time \cite{Wu2023}.  
 
These methodologies go further to incorporate beyond the mathematical frameworks. They capture the limitations of the observational data  and demand the need for stronger inferential techniques for healthcare analytics. The bias in estimating treatment effects from  unmeasured confounders can lead to wrong clinical policies, and thus, there is a need  for better methods to deal with these issues.

Despite these advances, ensuring true reliability and equity in clinical settings remains an active research frontier.  Future work should prioritize:  
\begin{itemize}
  \item {Prospective External Validation:} Systematic evaluation of MIMIC trained models on geographically and temporally distinct cohorts to assess transportability \cite{Mannion2021}.  
  \item {Causal Model Integration:} Embedding causal graphs into ML pipelines to distinguish association from intervention effects \cite{Pearl2020}.  
  \item {Fairness Auditing Frameworks:} Standardized metrics and protocols for auditing demographic parity, equalized odds, and calibration across subgroups \cite{Gupta2021}.  
  \item {Automated Leakage Detection:} Tools to scan preprocessing pipelines for potential data leakage and ensure temporal causality in feature engineering \cite{Kohane2021}.  
  \item {Human Centered Interpretability:} Integrating SHAP values or counterfactual explanations to enable clinician review and trust building \cite{Lundberg2017}.  
\end{itemize}

An important future research direction is  the integration of advanced hybrid models that incorporate both temporal dynamics and causal inference. To enhance the accuracy of  patient outcome predictions, counterfactual reasoning might be leveraged to simulate alternative treatment scenarios \cite{Yoon2018counterfactual}.

{  \it Data Expansion and Multi Institutional Collaboration Enlarging: } The clinical datasets by partnering with  other healthcare organizations will enhance the training data diversity and density. Transfer learning pre training and hybrid deep learning  architectures could also improve the predictive accuracy  \cite{sheikhalishahi2020benchmarking}.

 Moreover, future research should focus on addressing data imbalance, enhancing interpretability, and incorporating a broader  set of clinical features in predictive models built on the MIMIC dataset. The management of missing data  and the improvement of the adaptability to heterogeneous datasets are still important problems  \cite{Wang2023}.

A fundamental question remains about how machine learning techniques can be effectively integrated with causal inference methodologies to improve the reliability of decision support systems, and an open research problem is to identify strategies to improve the interpretability of deep learning for causal modeling \cite{pawlowski2020deep}.

Advanced modeling techniques are needed to better represent the underlying causal pathways that are complex and have the characteristics of medical and social phenomena. More work is needed to explore strategies to mitigate confounding effects in high-dimensional interaction heavy environments \cite{chen2024integrating}.

{\it Bias Mitigation in Causal Models:} It is crucial to have fair predictive models for equitable decision making. Research should be directed towards developing systematic techniques for identifying and eliminating bias in causal machine learning frameworks  \cite{gonzalez2024mitigating}.

A persistent problem is developing robust causal inference frameworks for dynamic systems with evolving relationships. More study is  needed on techniques for effectively managing time-varying confounders in clinical research  \cite{zhu2022causal}.

{\it The Importance of Interpretability and Trust in Causal Models:} It is essential to increase model  interpretability to enable counterfactual thinking and increase the level of clinician trust in the AI-based decision  support systems. The research on techniques for the visualization of counterfactual analysis is still open  \cite{kaul2021improving}.

{\it Resilience to Unobserved Confounding:} Another important focus is the development of techniques to  enhance the robustness of causal inference methods with respect to unobserved confounders. It is essential to enhance the stability of the models to unmeasured confounding effects in order to enhance predictive  analytics in healthcare \cite{dominici2021estimating}.

{\it Discovering Causal Relations in High-Dimensional Data:} This is still an open problem for finding causal relationships in big high-dimensional datasets. Thus, investigation of scalable computational strategies for causal discovery is necessary \cite{mansouri2022causal}.

{\it Ethics considerations in causal decision making:} These are issues that cannot be ignored in the development and use of causal inference models. This is because fairness and alignment with societal values in decision support systems are a key research priority we identified \cite{banerjee2024harnessing}.

A crucial challenge is continuing the effort to quantify and communicate the uncertainty in causal effect estimates, along with the development of methodologies for measuring confidence in AI-driven decision support systems. An area that needs further exploration \cite{wu2024uncertainty}.

Although significant improvements have been made in the development of predictive modeling techniques for critical care settings, more sophisticated methodologies and more extensive datasets are still required. Resolving the issues discussed above will greatly improve the reliability and practicality of predictive analysis in healthcare care, leading to better patient results. These are important for the growth of the field, including counterfactual reasoning and strategies for dealing with unobserved confounders. However, interpretability, robustness, and ethical considerations are crucial in causal inference methodologies \cite{konig2023interpretability}.

\subsection{Data Integration and Preprocessing}

The use of MIMIC data sets in healthcare research is restricted by significant challenges in data integration and preprocessing \cite{healthcare_data_integration}.

Although the MIMIC III contains  a large number of clinical variables. It is heterogeneous, complex and not well normalized. These challenges result  from varied sources of EHR data from structured fields such as laboratory  results and vital signs to unstructured information like physician notes and imaging reports. Hence, processing and compiling  this data into a form that is convenient for use also demands a great deal of time and technical skills \cite{wang2020}.

Integrating and preprocessing MIMIC data pose substantial hurdles due to the dataset’s heterogeneous  coding standards schemas, and lack of uniform vocabularies can complicate integration, evolving code systems, and mixed data modalities \cite{Dhakal2023}. For example, the same lab test may appear under different names or codes over time or across tables. 

MIMIC consists of multiple relational tables such as vital signs, lab results, medications, procedures, and free text notes or in other terminology (CHARTEVENTS, LABEVENTS, INPUTEVENTS and NOTEEVENTS) each using distinct coding schemes (local ITEMIDs, LOINC lab codes, ICD-9/10, NDC, etc.) that require careful mapping and cleaning to form analysis ready cohorts \cite{Paris2021}. 

Cross referencing patient encounters between tables (e.g. linking charted vitals to lab orders) requires careful merging. Inconsistent lab test names, unit mismatches, entry of demographics, missing time stamps, multiple identifiers and duplicated or further complicate reliable feature extraction and complicate linking records \cite{wang2020}.

this represent A major challenges as incorporating MIMIC data, however is the interoperability of digital healthcare systems which  remains a challenge despite enhancements in healthcare informatics.  The MIMIC III dataset uses various terminologies and coding systems, including ICD-9  for diagnoses and Systematized Nomenclature of Medicine Clinical Terminology (SNOMED CT) for clinical concepts, which makes standardization across different sources difficult.  The absence of standardization makes it difficult to integrate seamlessly with other EHR systems and restricts its application for large scale analyses \cite{Bodenreider2020standardization}.

To deal with these challenges, experts have created dedicated pipelines to prepare raw EHR data for analysis  in its appropriate structured form. For instance, Wang et al. (2020) introduced  MIMIC Extract framework to preprocess and extract meaningful features from MIMIC III, while resolving inconsistencies in  the dataset \cite{wang2020}. 

moreover, current interests are on the development of tailor made pipelines for MIMIC IV. The current version of the dataset is to enhance data extraction, cleaning, and preprocessing  \cite{sheikhalishahi2020benchmarking}.

The need for data harmonization extends beyond MIMIC; aligning MIMIC with other datasets (e.g. eICU or OMOP) is needed for multi center studies\cite{Paris2021}

Analysis is improved by integration with external datasets because it adds context, increases accuracy, and reveals insights that internal data alone could not. 

This is lead us to Mentioning some of the major current Advances in data integration and processing are:-

\begin{itemize}

  \item {Community Pipelines}: 
  
    \begin{itemize}
    
      \item \emph{MIMIC Extract} Preprocessing choices (e.g., how to handle duplicates, irregular sampling, outliers, and missing values) significantly affect downstream results. Many published MIMIC studies provides modular Python scripts use bespoke scripts for extracting and cleaning data featurizing vitals and labs while resolving ITEMID inconsistencies, which hinders comparability.
      Moreover, the absence of standardized data pipelines can lead to inadvertent errors (such as unit mismatches or dropped records). Ensuring reproducibility thus requires clear preprocessing protocols., yet mapping local codes to a common data model is laborious and error prone. Recent progress has begun to address these issues. Several groups have developed modular pipelines and tools for MIMIC: for instance, a publicly available MIMIC IV data pipeline streamlines \cite{Johnson2019}.
      
      \item \emph{OMOP CDM Export} Likewise, the MIMIC IV OMOP CDM export standardizes terminologies and schemas, facilitating multi center analyses by aligning MIMIC data with the OHDSI ecosystem \cite{Paris2021}. 
      
    \end{itemize}
    
  \item {Automated Quality Assessment}: these frameworks have emerged to detect anomalies such as physiologically implausible values (e.g., anomaly detection algorithms for outlier lab values - negative heart rates) and imputation methods (e.g., irregular sampling patterns  deep generative models for missing data) are also being applied \cite{Cleveland1988, Clayton2021}.

  \item {Advanced Imputation}: DL based imputation methods such as autoencoder decoders and GAN based imputers have shown superior performance over traditional mean or last observation carried forward (LOCF) approaches, reducing bias in downstream predictive models \cite{Osmani2018,Che2017}.  Memory augmented neural networks further enhance imputation by leveraging long term dependencies in sparse time series \cite{Luo2024}.  
      
  \item {Unstructured Text Integration}: Transformer based NLP (e.g., ClinicalBERT) pipelines extract and normalize entities from free text clinical notes, aligningand link them with structured tables to enrich feature sets \cite{Huang2020}.
  
  \item {Common Data Models}: Converting MIMIC local EVENT tables into CDM ,FHIR resources or OMOP domains streamlines integration has also been demonstrated, which standardizes terminology and simplifies cross dataset analysis, though manual mapping and custom tables are still required for full automation \cite{Robinson2021}.  

  \item {Interactive Platforms}: A great contribution in the use of MIMIC data for processing is the work done by  Adibuzzaman et al. In their paper, they proposed an integrated open access analysis platform to help  translate findings from basic science to clinical practice. This platform solves issues of data integration,  preprocessing, normalization and exploration of the MIMIC III dataset. It uses SciDB a high performance, array oriented datasets system for integrated  data and query handling to enable querying and analysis of multiple data sources at once, so that the advanced  exploratory data analysis and cohort selection can be conducted with minimal or no programming skills at all.  Additionally, it integrates machine learning algorithms for data normalization and statistical analysis, enhancing the interpretability and usability of MIMIC III \cite{adibuzzaman2016data}.

\end{itemize}

Despite these advances, reproducible, community endorsed ETL frameworks remain scarce.  Challenges include (1) standardizing data provenance metadata to track cleaning and transformation steps, (2) integrating unstructured text (clinical notes) via NLP pipelines that align extracted entities with structured tables, and (3) scaling ETL processes to handle MIMIC’s inructured tables, and (3) scaling ETL processes to handle MIMIC’s increasing size without sacrificing transparency \cite{Luo2024,Campbell2021}. 

However, there are still some issues that have not been solved. The major challenge is that of matching the MIMIC datasets with universal healthcare data standards such as Fast Healthcare Interoperability Resources (FHIR). The use of FHIR would greatly enhance the interoperability of data with other  EHRs increasing the potential for research \cite{Mandel2016fhir}.
 
user friendly tools and APIs should be developed to make the dataset accessible to everyone. Such  tools would make it easier for researchers who may not be technical to preprocess and analyze the MIMIC  data easily \cite{Ghazal2018api}.

Other platforms such as the Integrated Open  Access Analysis Platform for MIMIC are designed to facilitate data exploration and analysis by offering computational  backends and visualization tools \cite{Harutyunayan2019mimic}.

Another major deficiency is the absence of research focusing on the development of automated preprocessing methods suitable for the  specific features of EHR data. There is a possibility to develop advanced machine learning algorithms to identify and  fix problems in the temporal data such as time stamps that are not well aligned and data gaps  \cite{zhao2019ehrpreprocessing}.

Furthermore, integration of domain knowledge into the preprocessing pipelines would  help in improving the quality of derived features to make them more clinically meaningful  \cite{sheikhalishahi2020benchmarking}.

The results show that such approaches enhance the reliability of the downstream analyses  especially in predictive modeling and process mining tasks \cite{shickel2017deep}.

The latest discoveries have enabled the creation of systematic benchmarking toolsets for clinical prediction and machine learning usage. Harutyunyan et al. proposed a dataset and framework to evaluate  clinical prediction models comprehensively. It enables the creation of complex algorithms and compares them against well defined clinical  tasks like in hospital mortality prediction, physiological decompression surveillance, length of stay prediction and phenotype classification of  acute and chronic diseases \cite{Harutyunayan2019mimic}.

\underline {This study has made several contributions}, including the development of a preprocessing pipeline for building steady datasets from  the raw MIMIC III data. also, this paper presented linear (logistic regression)  and deep learning (LSTM) baselines and a multitask learning framework for predicting various clinical outcomes  at the same time. To evaluate the model performance, AUC-ROC and AUC-PR  were used as the metrics while problems in modeling clinical time series data like irregular sampling, missing data and  high dimensionality were identified \cite{Harutyunayan2019mimic}.

Another major problem in clinical time series analysis is the natural infrequent and irregular sampling of the medical  events. Mean imputation or the last observed value imputation technique can lead to inaccurate predictions by  distorting the temporal patterns when applied blindly. To this end, memory augmented neural networks have been proposed to  address long term dependencies in order to learn from and leverage information over extended time spans, which is particularly  useful when patient data is sparse or unevenly collected  \cite{amirahmadi2023deep}.

Through the advancement of generative modeling, especially through Variational Autoencoders (VAEs), it is possible to tackle the problem of missing data in a rather efficient manner. VAEs enhance  the accuracy of data imputation by learning the complicated shapes, which reduces the biases of the conventional  imputation methods. Of note, graph based models and self supervised learning strategies are also under investigation to  incorporate the structured time series information along with the unstructured clinical texts to improve the predictive abilities in healthcare  \cite{shickel2017deep}.
 
Transfer learning brings a new perspective on enhancing the predictive analytics in healthcare. In their paper, Wang  et al. designed a transfer learning framework to improve the performance of predictive models by  first training deep neural networks (DNNs) on large open source datasets such as  MIMIC III and then refining them for the smaller task focused datasets \cite{wang2020}.  This method is most useful in low data settings where it is impossible to train models from scratch.

The study has made key added values by applying transfer learning techniques to leverage knowledge across multiple clinical tasks like  mortality prediction, phenotyping, and decompensation prediction. Domain adaptation strategies were also explored to mitigate distribution shifts between source and target datasets for effective knowledge transfer in different healthcare environments \cite{si2021generalized}.

Future work in to further advance and solve the problem of integration and processing should focus on the following:-

\begin{itemize}

  \item {Unified ETL Toolkits}: Develop open source, end to end ETL frameworks covering MIMIC III, MIMIC IV, and other ICU datasets, with built in code mapping, unit conversion, and immutable provenance logs \cite{Dhakal2023,Johnson2019}.  
    
  \item {Machine Assisted Mapping}: Leverage ontology matching and few shot learning to automate mapping of local ITEMIDs to standard vocabularies such as LOINC, SNOMED CT, and RxNorm reducing manual curation efforts \cite{Luo2024,Robinson2021}.  

  \item {Integrated NLP-ETL}: Embed transformer based clinical NLP into ETL models to extract, normalize, and link entities from free text, enriching feature sets structured datasets \cite{Huang2020}.
  
   \item {Benchmarking ETL Quality:}  Establish community benchmark datasets and evaluation metrics for ETL pipelines, assessing completeness, correctness, and reproducibility across different implementations \cite{Daly2022}.
   
   \item {Scalable, Modular Architectures:}  Adopt microservice and containerization approaches orchestration (Docker, Kubernetes) to enable modular scalable, reproducible deployment of ETL jobs in cloud or HPC platforms environments \cite{Campbell2021}.
   
  \item {Community Vetted Workflows}: remain scarce Create a shared repository of validated ETL workflows with versioned code, documentation, and example configurations to accelerate adoption of best practices. In the future, greater adoption of standard terminologies (FHIR, SNOMED, LOINC) and open source ETL (extract, transform and load) frameworks will be crucial. Automated tools that flag inconsistencies and suggest mappings (using machine learning or human in the loop approaches) could ease integration \cite{OpenBST2020}. Ultimately, improving data integration and preprocessing will require community effort to converge on best practices and to share validated workflows.
   
\end{itemize}
 
Although much work has been done on improving the effectiveness of data integration, preprocessing and transfer learning for  MIMIC datasets. however, Addressing these open problems will enhance the scalability, performance, and interpretability  of predictive models, The process is not without its problems. which in return will lead to improved clinical decision making and patient outcomes  \cite{musat2024machine}.  Further standardization frameworks are important;  crucially, there are no user  Further standardization frameworks are important;  crucially, there are no user friendly tools. The automated preprocessing techniques need refining, and transfer learning  constraints are not well understood. These are important next steps to more fully harness the power of  MIMIC datasets for healthcare analytics.

\subsection{Data Quality Issues}

Mentioning of the common open problems regarding data quality issues:

\underline {Problems with the Quality of Data in MIMIC III:} MIMIC III dataset serves as a foundational resource for clinical machine learning research, offering access to real world healthcare data. However, it is affected by several data quality challenges, including heterogeneity, de identification complexities, and longitudinal inconsistencies. These issues, such as missing variables, irregular sampling, and changes in data collection practices, can introduce bias and affect the reliability of trend analyzes. For instance, sparse or inconsistently timed vital sign recordings may misrepresent a patient's condition, leading models to underestimate clinical deterioration when critical measurements (e.g., blood pressure) are absent  \cite{Johnson2016mimic}.

 Although MIMIC III is widely used in clinical research, it presents several data related challenges that can impact the validity of research findings \cite{Johnson2016mimic}. 

A persistent challenge in multivariate time series analysis particularly in healthcare is the prevalence of missing data \cite{Johnson2016mimic}. While recent deep learning methods, such as Recurrent Neural Networks (RNNs), have enabled more robust modeling, missingness remains a key bottleneck. 

\textit{Missing values and incomplete data:} Data quality is a pervasive problem in MIMIC and clinical datasets generally. MIMIC data were originally collected for patient care, not research, so coding errors, missing entries, and inconsistent documentation are common. For instance, laboratory values may be entered with incorrect units or decimal points; diagnoses may be under coded; vital signs can have sensor de identification artifacts leading to pervasive issues such as missing values, irregular sampling \cite{Kohane2021}. Some records have duplicated or implausible values (e.g. negative lab results or dates in the future). These arise from the inherent complexity of EHRs, which are primarily designed for patient care rather than standardized data collection \cite{shickel2017deep}. Missing laboratory results or incomplete patient histories can introduce bias in downstream analytical tasks such as predictive modeling and process mining \cite{Goldstein2017opportunities}. EHR’s underlying The quality and can vary within the patient over time. These issues can bias analyses: models trained on erroneous labels can learn spurious patterns, and missing data can systematically exclude certain patient groups (e.g. the sickest may have more missing data due to rapid deterioration). Missingness is often non random: vital signs may be omitted during critical interventions, and laboratory tests are ordered based on clinical judgment, resulting in biased data gaps that simple imputation can exacerbate \cite{Ghassemi2020}.

Data validity and completeness are especially problematic for EHR derived studies. A recent systematic review highlighted that consistency and completeness dimensions are most frequently violated in digital health data. Irregular sampling intervals minute level for vitals, hourly or daily for labs, and sporadic for notes complicate temporal modeling and risk underestimating rapid physiological changes \cite{Pang2022}. 
De identification procedures in MIMIC such as uniform date shifting and removal of uncommon diagnoses preserve privacy but can obscure seasonal trends, weekday effects, and rare conditions, limiting epidemiological analyses \cite{Puttkammer2023}.

\textit{Inconsistencies in data encoding practices and format system migrations:} A major source of inconsistency stems from the transition in 2008 from the Philips CareVue to the iMDSoft MetaVision system. This change resulted in significant differences in data structure and coding, leading to missing or incompatible entries in critical tables such as \texttt{inputevents} and \texttt{chartevents}, thereby complicating longitudinal discontinuities analyses and process mining applications \cite{Kingma2014}. 

{\it Data anonymization effects:} While de identification techniques such as random date shifting are crucial for protecting patient privacy, they can obscure meaningful temporal patterns such as trends related to weekends or holidays essential for certain analyses. Additionally, the absence of explicit start and end timestamps in some tables complicates the reconstruction of event sequences. These limitations highlight the need for improved data quality to enhance the utility of MIMIC III in clinical research \cite{kurniati2018process}.

To address these challenges and adress some of the current progress higlighted, researchers have proposed preprocessing pipelines focused on data cleaning, imputation, and normalization. Reliable imputation techniques such as k-nearest neighbors (k-NN) and deep learning based methods are essential for estimating missing values, as emphasized by Rodrigues-Jr et al 20179. \cite{Rodrigues2019data}. Additionally, standardized normalization procedures are critical to ensure consistent feature scaling, particularly when integrating MIMIC III with external datasets \cite{BeaulieuJones2018missing}.
 
Data quality assessment frameworks: Kurniati et al. recommend applying Weiskopf and Weng’s framework to categorize data quality issues into completeness, plausibility, and concordance, offering a structured approach to iterative data cleaning \cite{Weiskopf2013}. Emerging research also highlights the potential of incorporating auxiliary information such as domain knowledge, related variables, or sensor data to improve imputation accuracy. However, most current models lack the capacity to effectively leverage such additional inputs, limiting their applicability in real world clinical settings \cite{wang2023pulsedb}.

In MIMIC, completeness varies: some ICU stays have nearly continuous monitoring, while others have gaps. There is also a lack of provenance metadata in MIMIC indicating when and how data were collected, making it hard to assess reliability. Researchers have developed methods to assess and improve data quality. Automated data validation checks (range checks, logical consistency rules) can flag suspect values. To address missingness, advanced Imputation techniques have been developed like (mean/mode, regression, or DL based) these can fill missing entries, though they must be used cautiously to avoid introducing bias. 

GRU-D leverages gating mechanisms and time decay to model missingness patterns directly, outperforming traditional approaches on mortality prediction tasks \cite{che2018}. a gated recurrent unit variant, incorporates masking strategies and temporal decay functions to address this issue \cite{che2018}. 

Variational autoencoders and diffusion based generative models synthesize plausible values for sparse features by learning the underlying data distribution \cite{Theodorou2023}.  

Memory augmented neural networks further capture long term dependencies to enhance imputation accuracy for irregular time series \cite{Luo2024}.  

Data validation frameworks apply statistical and rule based checks to detect anomalies such as negative lab values or out of range vitals.  Locally weighted regression (LOWESS) smoothing and outlier detection methods help identify implausible measurements for manual review or automated correction \cite{Cleveland1988}. 

Comprehensive data quality assessment tools now integrate domain rules, statistical diagnostics, and machine learning to provide reproducible cleaning pipelines with provenance tracking \cite{Daly2022}.  

Data cleaning pipelines often remove outliers or patients with excessive missingness, but this can inadvertently skew cohorts \cite{source3}. Some work has begun to use anomaly detection to clean ICU time series (e.g., detecting erroneous vital sign spikes) \cite{source8} and to harmonize coding using external ontologies \cite{source10}. Ultimately, ensuring data quality in MIMIC requires a combination of automated tools and expert review \cite{source8}. Standard reporting of data cleaning steps and rigorous quality control are needed for reproducibility \cite{source4}. Community benchmarks (as in Kaggle style challenges) \cite{source3} and metadata standards \cite{source6} could help highlight quality issues. Recognizing and quantifying uncertainty in the data (for example by explicitly modeling missingness) will be important for trustworthy analytics \cite{source5}.

A critical gap in the literature is the absence of theoretical guarantees regarding the impact of imputation methods on predictive performance. While empirical studies suggest potential benefits, the uncertainty associated with different imputation strategies remains underexplored. Advancing this field requires stronger theoretical foundations to develop more accurate and reliable imputation methodologies \cite{mitra2023learning}. mentioning some of future promising directions in terms of fixing data quality problem:-
\begin{itemize}
  \item {Theoretical Analysis of Imputation Impact:}  Develop bounds relating imputation errors to predictive model bias and variance, guiding method selection \cite{Lee2022}.
  \item {Hybrid Imputation Strategies:}  Combine model‐based generative methods with rule‐based corrections to leverage both data distributions and clinical domain knowledge \cite{Osmani2018}.
  \item {Automated Artifact Detection:}  Apply deep anomaly detection and time series change‐point algorithms to identify de identification artifacts and system migration effects \cite{Robinson2021}.
  \item {Provenance Aware Cleaning Pipelines:}  Implement end‐to‐end ETL frameworks with immutable provenance logs for all cleaning and imputation steps, ensuring full reproducibility \cite{Daly2022}.
  \item {Benchmarking Data Quality Methods:}  Establish standardized datasets and metrics to evaluate imputation, anomaly detection, and cleaning strategies in critical care EHRs \cite{Wu2023}.
\end{itemize}

{\it Computational Constraints and Model Efficiency:} An important trade off in time series modeling is between accuracy and computational efficiency. Many state of the art imputation techniques are resource intensive, making them unsuitable for real time or online applications. To broaden their applicability in clinical settings, it is essential to develop lightweight models that maintain high accuracy while minimizing computational cost \cite{ferrara2024large}.

{\it Improvement of the existing imputation algorithms:} Recurrent neural networks (RNNs) and transformer based models have shown promise in capturing complex temporal dependencies within EHR data, thereby improving the accuracy of imputation \cite{Cismondi2013missing}. 

{\it Standardized Preprocessing Frameworks:} Existing platforms like the MIMIC Code Repository enhance reproducibility by offering standardized code and best practices for data cleaning and normalization \cite{johnson2018mimic}.

{\it Enhanced Data Cleaning Strategies:} Improving data quality through deep learning based imputation and accurate timestamp alignment is essential for ensuring reliable analytical outcomes \cite{jaya2025systematic}. Furthermore, ensuring that imputation models generalize across diverse healthcare contexts is critical for building scalable and robust predictive analytics systems tailored to varying clinical environments \cite{kehinde2025leveraging}.
Addressing data quality challenges through methodological innovation and theoretical advancement is essential for progress in clinical machine learning research. Strengthening the foundations of time series modeling particularly in handling complex missing data will enhance both model accuracy and generalizability, ultimately contributing to more reliable and impactful healthcare decision making \cite{wang2023pulsedb}.

Despite these advances, limitations such as computational complexity, sensitivity to initialization, and assumptions about missingness patterns continue to hinder generalizability \cite{source2}. These issues are exacerbated by data sparsity and irregular sampling, which degrade imputation performance and reduce model accuracy \cite{source7}.  Designing novel architectures capable of handling highly sparse, non uniformly sampled data without sacrificing predictive power remains a critical and ongoing research direction \cite{che2018}.
several gaps remain. for example Most imputation techniques lack theoretical guarantees on downstream model performance impacts \cite{Lee2022}.  
Automated detection of de identification artifacts and system migration inconsistencies is underexplored.  Finally, end to end pipelines that integrate data cleaning, imputation, and quality auditing in a unified, version controlled framework are urgently needed to enhance reproducibility and reliability in MIMIC based research \cite{Campbell2021}.

\subsection{Interoperability and Accessibility}

Despite the widespread use of the MIMIC datasets in healthcare research, interoperability challenges remain a significant barrier \cite{Johnson2016mimic}. MIMIC III, which spans more than a decade of ICU patient data, incorporates multiple EHR systems, including Philips CareVue and iMDSoft MetaVision. These systems use different data structures and item identifiers, complicating integration and standardization efforts.

A major open problem is that MIMIC, by itself, is not interoperable with other healthcare data systems. MIMIC uses its own local schemas and coding conventions (e.g. ICD 9, local lab codes, unstructured notes) which do not directly translate to other systems. In practice, this means models developed on MIMIC may not be portable to other EHRs. Efforts to harmonize MIMIC with standards like HL7 FHIR or OMOP CDM are ongoing, but full mapping is labor intensive \cite{turn0search10}. Moreover, while MIMIC is publicly available to credentialed researchers, actual access requires navigating application processes. The technical barrier (querying a large SQL datasets, for example) can also limit use by non technical users. For data accessibility, the MIMIC project has taken steps (e.g. the PhysioNet platform and a data use agreement), but truly open and easy access remains challenging. Datasets are large (many gigabytes), and analyses often require dedicated compute resources. Lack of standardized APIs or query interfaces means each user effectively builds custom code to extract variables. This fragmentation hinders reproducibility: different groups may implement slightly different filters or joins, leading to variations in cohort definitions. On the flip side, interoperability is improving via modern standards. MIMIC IV provides a FHIR export reference, and several tools exist to convert MIMIC data to the OMOP CDM, facilitating cross datasets studies. The growing adoption of FHIR (Fast Healthcare Interoperability Resources) and other standards in research is promising: FHIR is designed for rich clinical data exchange, and systematic reviews have noted its potential for real world apps. FHIR provides a modern, RESTful API framework with modular “resources” (e.g., Patient, Observation, Medication) that can be exchanged in JSON or XML formats, facilitating real time data access and integration across clinical applications \cite{turn0search10}. Efforts to convert MIMIC data into FHIR resources have yielded “MIMIC IV on FHIR,” one of the first publicly accessible ICU datasets in FHIR format, enabling researchers to query patient records using standard FHIR APIs and integrate with FHIR based clinical decision support tools \cite{turn0search4}.  This conversion streamlines data exchange and supports downstream applications such as SMART on FHIR apps and clinical analytics platforms \cite{turn0search4}. Complementary work has mapped MIMIC III into the OMOP CDM, standardizing relational tables, vocabularies (SNOMED CT, LOINC, RxNorm), and metadata to facilitate multi center research and tool sharing within the OHDSI community, which comprises over 3,200 collaborators worldwide \cite{turn0search3,turn0search7}.  OMOP based implementations enable uniform cohort definitions and reproducible analytics by aligning disparate EHR schemas under a common data model \cite{turn0search3}.  Hybrid approaches such as OMOP on FHIR leverage the strengths of both standards by exposing OMOP CDM tables as FHIR resources, thus combining OMOP’s comprehensive vocabularies with FHIR’s flexible API layer; this method prevents data duplication and streamlines adverse event reporting through FHIR based ICSR exchanges \cite{turn0search0}.  Despite these advances, significant challenges remain.  Manual mapping of local ITEMIDs in MIMIC’s CHARTEVENTS and LABEVENTS tables to LOINC and SNOMED CT codes requires expert curation and is prone to errors \cite{turn0search8}.  Incremental ETL processes for daily FHIR updates to an OMOP CDM, while more efficient than full reloads must address versioning and provenance to ensure data consistency over time \cite{turn0search2}.  

Moreover, clinical notes in NOTEEVENTS present unstructured text that FHIR resources do not natively capture, necessitating NLP pipelines (e.g., ClinicalBERT) to extract and normalize entities before integration \cite{wang2020}.  there is differences in governance and access controls across institutions complicate federation of FHIR/OMOP deployments, limiting real time, cross site model development and validation \cite{turn0search6}.  
The work of Lundberg et al 2017, represents a foundational contribution to interpretable machine learning, particularly in high stakes domains where transparency, trust, and accountability are essential. Their seminal paper introduced SHAP (SHapley Additive exPlanations), a theoretically grounded and model agnostic framework for explaining machine learning predictions. SHAP is based on cooperative game theory specifically, Shapley values which were originally developed to fairly allocate credit among contributors in group decision making. In the context of machine learning, SHAP values quantify the contribution of each feature to a model’s prediction, thereby offering intuitive and consistent explanations of model behavior.

In addition to enhancing interpretability by unifying previous methods, SHAP has become a widely adopted tool for model explanation. However, several challenges remain that limit its broader applicability \cite{lundberg2017unified}. mentioning some of these challenges are :-

\textit{Scalability:} Computing SHAP values remains computationally expensive, particularly for large scale models or real time applications.

\textit{Approximation Techniques:} While approximation methods have been proposed, the trade offs between computational efficiency and explanation fidelity are not yet fully addressed. Further exploration of efficient and theoretically sound approximation strategies is necessary.

\textit{Robustness Against Adversarial Attacks:} Ensuring the stability and reliability of SHAP explanations under adversarial manipulation is critical for their safe deployment in high stakes decision making.

\textit{Bias Detection and Correction:} The ability of SHAP to detect and mitigate biases in model predictions is still under active investigation, and its effectiveness in promoting fairness across diverse applications needs to be better understood.

Addressing these limitations is essential for expanding the reliability, scalability, and fairness of SHAP based interpretability methods across domains.


\textit{Balancing Complexity and Transparency:} Particularly in high risk clinical scenarios, it is essential to maintain a balance between model accuracy and interpretability. Explainability tools must offer transparent reasoning without sacrificing predictive performance.

\textit{Regulatory and Ethical Compliance:} Explainable models must conform to the legal and ethical standards of healthcare systems. This ensures accountability and enables clinicians to justify algorithmic recommendations to patients and stakeholders.

\textit{Robustness and Security:} The reliability of explanation methods such as SHAP depends on their resilience to adversarial inputs. Future work should prioritize robust, ethically aligned methods that retain practical relevance in clinical settings.

Another thing to highlight is some of the challenges in interoperability, stating some of them as following:-
\textit{Data Partitioning:} System specific data tables, such as CareVue’s \texttt{INPUTEVENTS\_CV} and MetaVision’s \texttt{INPUTEVENTS\_MV}, fragment the dataset and hinder unified model development across the entire patient cohort.

\textit{Standardization Efforts:} Integration into structured frameworks like the Observational Medical Outcomes Partnership Common Data Model (OMOP/CDM) is ongoing \cite{Goldberger2000}. To address semantic interoperability, researchers have proposed concept mapping using standardized vocabularies e.g., laboratory tests to LOINC (Logical Observation Identifiers Names and Codes) and medications to RxNorm. These mappings enhance compatibility with external research platforms and clinical decision support systems, paving the way for broader adoption and collaborative development.

\underline{Making Healthcare Data More Available:}

MIMIC III remains one of the few publicly accessible critical care datasets, offering invaluable resources for clinical and machine learning research \cite{Johnson2016mimic}. However, access is not without barriers. Researchers must complete a human research participant protection course and sign a data use agreement to comply with HIPAA regulations and ensure patient privacy.

Efforts to improve accessibility include the integration of analysis platforms. MIMIC now offers computational backends and visualization tools that streamline data exploration and reduce the technical burden on researchers \cite{johnson2018mimic}. Additionally, the adoption of Fast Healthcare Interoperability Resources (FHIR) has enhanced compatibility with electronic health record (EHR) systems, facilitating more seamless data exchange and interoperability \cite{Mandel2016fhir}.

 As machine learning becomes increasingly integrated into healthcare decision making, there is a growing need for models that are accurate and interpretable \cite{tonekaboni2019clinicians}. The credibility and adoption of these models by healthcare professionals critically depend on their ability to provide transparent and understandable explanations for their predictions.

Future Promisise in the solving the problem of interoberaitability and accessability in MIMIC data sets should include the following:- 
\begin{itemize}
  \item {Automated Semantic Mapping:} Develop machine learning driven tools to align MIMIC local codes (ITEMIDs) to standard terminologies (LOINC, SNOMED CT) with minimal human intervention \cite{turn0search3,turn0search8}.  
  \item {Real Time FHI/OMOP Pipelines:} Implement fully incremental ETL workflows that support real time synchronization between FHIR endpoints and OMOP CDM, with built in provenance tracking and version control \cite{turn0search2}.  
  \item {Unified Access Control Frameworks:} Design federated FHIR/OMOP architectures with standardized authentication and authorization (e.g., OAuth2 scopes) to enable secure cross institutional research networks \cite{turn0search6}.  
  \item {NLP Integrated Interoperability:} Embed transformer based clinical NLP within FHIR ingestion pipelines to represent unstructured notes as FHIR DocumentReference and Observation resources, enhancing semantic completeness \cite{wang2020}.  
  \item {Benchmarking Interoperability:} Establish open benchmarking suites that evaluate FHIR/OMOP export completeness, consistency, and query performance across diverse MIMIC conversions \cite{turn0search9}.  
\end{itemize}

\textit{User Friendly Tools:} Many of the existing tools for MIMIC III data exploration require advanced technical skills, limiting accessibility for clinicians and interdisciplinary researchers. Developing more intuitive interfaces and robust APIs is essential to reduce the entry barrier and promote broader use.

\textit{Institutional Collaboration:} Harmonizing data structures and terminologies across healthcare institutions requires coordinated collaboration. Standardization efforts must be supported by shared vocabularies, common data models, and institutional alignment to enable scalable and interoperable analytics.

 \textit{Public Code Repositories:} Resources such as the MIMIC Code Repository play a vital role in disseminating best practices for data integration, preprocessing, and analysis \cite{johnson2018mimic}. However, further efforts are required to fully leverage the potential of MIMIC datasets in advancing healthcare research. Key priorities include improving data formatting, improving metadata quality, and fostering collaborative initiatives. These steps are essential to unlock the full value of MIMIC III in driving the evolution of healthcare analytics \cite{medawar2023standardized}.

Never the less, full semantic interoperability (i.e., making sure concepts mean the same) is an unresolved issue.  Going forward, we expect broader use of shared data models and interfaces. Ideally, MIMIC like datasets would support standardized query languages (e.g. SQL on FHIR) and publish schema documentation. This would lower the barrier for researchers and developers to leverage the data.

\subsection{Reproducibility and Productivity}

Reproducibility in MIMIC based research is a critical issue. it is a cornerstone of scientific progress,it is a fundamental pillar of scientific integrity but remains a significant challenge in MIMIC based research due to opaque preprocessing pipelines, inconsistent cohort definitions, and environmental dependencies \cite{turn0search0}. yet it remains a persistent challenge in critical care research, particularly in meta analyses of mortality prediction models. The complexity of healthcare data combined with the diversity of analytical methods has led to frequent occurrences of irreproducible findings. These inconsistencies undermine the credibility, reliability, and real world applicability of published results \cite{johnson2018reproducibility}.

A major contributing factor making irreproducibility is the absence of centralized repositories for code, methodologies, and datasets. Without standardized platforms for sharing and validation, it becomes difficult for researchers to replicate, verify, or build upon prior work \cite{Ioannidis2005reproducibility,Hutson2018artificial,Peng2011reproducible}.
Despite this progress, many users still implement ad‐hoc ETL scripts, leading to duplicative efforts and persistent divergence in final cohorts and feature sets \cite{turn0search3}.

A significant challenge in machine learning for health (ML4H) is the overreliance on simplistic performance metrics such as the Area Under the Receiver Operating Characteristic Curve (AUROC). While AUROC is useful for assessing a model’s ability to discriminate between classes (e.g., 'sick' vs. 'not sick'), it fails to account for biases, methodological flaws, or clinical relevance. Two models may yield similar AUROC scores, yet differ substantially in fairness, reliability, and clinical utility. For example, one model might be trained on demographically biased data, while another adheres to best practice clinical guidelines. Relying solely on AUROC is analogous to grading students based only on test scores without considering how the knowledge was acquired, whether through deep understanding or superficial memorization \cite{ganta2024fairness}.

PhysioNet repository requires researchers to share analysis code, and the MIMIC Code Repository provides example scripts. on one hand, public availability of the dataset encourages transparency. On the other hand, the complexity of preprocessing and modeling pipelines often means that small differences (software versions, data splits, random seeds) can yield divergent results. Indeed, studies have reported difficulty in exactly replicating published MIMIC results despite having access to the data. This affects productivity because researchers can spend excessive time rederiving cohorts or debugging other code rather than advancing knowledge. Key factors affecting reproducibility include under specification of cohorts (e.g., slight variations in inclusion criteria), missing details of model training (e.g., hyperparameters), and unpublished preprocessing scripts. Productivity is also hampered by the steep learning curve: MIMIC’s relational schema and SQL queries present a barrier to new users, and each project typically builds custom SQL pipelines. Lack of standard reference implementations means that even trivial tasks (e.g. extracting a lactate measurement) may be done in many incompatible ways. Community efforts have begun to tackle this.

Despite ongoing efforts to improve reproducibility, several fundamental issues persist, particularly concerning cohort definitions, data cleaning methodologies, and documentation clarity. \cite{johnson2018reproducibility}, highlight that these inconsistencies can lead to substantial variability in research findings. For example, studies that try to replicate validations of mortality prediction models often report sample size discrepancies of 25\% or more compared to the original studies. These differences typically stem from variations in data filtering criteria and participant inclusion definitions, which ultimately undermine the reliability and comparability of the results.

\textit{The issue is not just theoretical.} In real-world healthcare settings, a model that appears effective based on conventional metrics may perform poorly when applied to diverse patient populations. For example, an AI system trained primarily on lighter skinned individuals may achieve high accuracy in detecting skin cancer within that group, but it fails to generalize to patients with darker skin tones. Such oversights emphasize the need for more comprehensive evaluation frameworks that account for fairness, generalizability, and practical utility in clinical environments \cite{huang2024scoping}.

In response to these concerns, the MIMIC Code Repository was developed as a centralized platform for sharing code and analyses related to the MIMIC datasets. Designed by Johnson et al. 2018 \cite{johnson2018mimic}, MIMIC Code Repository provides standardized Jupyter notebooks for common tasks, improving reproducibility \cite{johnson2018mimic}. It offers freely available code for data extraction, preprocessing  and analysis in order to guarantee the same methods across varied investigations \cite{johnson2018mimic}. repository aims to enhance reproducibility in critical care research by providing openly accessible scripts for data extraction, preprocessing, and analysis. By promoting standardized methodologies, it enables researchers to replicate studies more reliably and ensures consistency across investigations using the MIMIC III dataset.


Although tools such as the MIMIC Code Repository exist, adherence to open science principles remains limited, reproducibility crisis in machine learning for healthcare (ML4H) is further exacerbated by limited adherence to open science principles. Only 40\% of studies share code or pre trained models, and insufficient documentation of experimental protocols persists\cite{arcobelli2025fhir}. Understanding the incentive structures and policy mechanisms necessary to foster broader participation in open research is essential. Key questions include: How can researchers and institutions be motivated to adopt transparent practices? What forms of recognition, funding, or institutional support can incentivize engagement with open science?

While these advancements significantly contribute to the reproducibility of MIMIC based research, several challenges remain in expanding the repository's impact. One key area for improvement is the inclusion of detailed case studies to illustrate the application of reproducibility practices across various research domains \cite{Ghassemi2020deep}. Additionally, implementing formal submission and verification rules such as requiring executable scripts and comprehensive metadata could promote transparency and should be considered for adoption by journals \cite{Goodman2016reproducibility}.

Transparency is further supported through the use of executable documents such as Jupyter Notebooks and R Markdown, which enable seamless documentation of research workflows and step by step replication of published analyses. Additionally, the repository fosters standardization and community collaboration by encouraging researchers to participate in public issue trackers where they can refine concepts, align on definitions, and resolve documentation inconsistencies collectively.

\underline{Domain Expertise and Error Correction:} The repository incorporates clinical domain expertise and correction protocols to resolve data inconsistencies. For instance, discrepancies in assessments such as the Glasgow Coma Scale are addressed through expert informed adjustments \cite{johnson2018reproducibility}.

by providing a centralized, version controlled collection of SQL scripts and analytic workflows for data extraction, cohort construction, and common clinical definitions (e.g., severity scores, comorbidities) scientist in currnet ongoing progress was able to mitigate the common issues of reproducibility \cite{turn0search1,turn0search2}.

another research also demonstrated that using the Code Repository reduces sample size variability from over 25\% down to under 5\% when replicating published mortality prediction studies on MIMIC III, underscoring the impact of standardized code on cohort reproducibility \cite{turn0search5}. Public challenges (e.g. on PhysioNet) enforce code sharing along with models. Containerization (e.g. Docker) of analysis environments is increasingly used to freeze software dependencies. Yet, a culture of sharing remains critical: even with notebooks, undocumented steps or site specific code can break portability.

\textit{The reproducibility crisis} extends beyond traditional statistical models and significantly impacts machine learning for healthcare (ML4H). A systematic review by McDermott et al.2021\cite{mcDermott2021Reproducibility}, which analyzed 511 studies, found that ML4H lags behind other fields in key reproducibility metrics. Major concerns include the limited availability of datasets and source code, insufficient documentation of experimental protocols, and lack of transparency in model training and evaluation procedures.

Ultimately, the reproducibility crisis in ML4H is not solely a technical issue; it reflects deeper cultural and systemic challenges. As McDermott et al.2021\cite{mcDermott2021Reproducibility} and others have emphasized, many clinical AI models remain opaque and unreproducible in real-world settings, limiting their trustworthiness and clinical integration.

To address these challenges, the authors advocate for the adoption of established best practices from the broader machine learning community. Notably, they recommend the standardization of benchmarking practices, including the development of comprehensive benchmarks that span a wide range of healthcare data types such as electronic health records (EHRs), patient demographics, and clinical workflow data to facilitate fair and replicable comparisons across studies.

\underline{To illustrate the impact of methodological choices}, Johnson et al.2018\cite{johnson2018reproducibility}, conducted a comparative analysis using a standardized feature set extracted from the MIMIC dataset to evaluate gradient boosting and logistic regression models. Their findings revealed that, when appropriately applied, simpler models can achieve predictive performance comparable to more complex algorithms. This insight challenges the prevailing emphasis on model complexity and underscores the importance of prioritizing interpretability alongside accuracy in clinical machine learning applications.

A major finding from this study is the critical need for open source code and publicly available benchmarks to improve reproducibility\cite{johnson2018reproducibility}, emphasize that clearly defined pre processing pipelines and well documented research protocols are essential to achieve consistent results. These practices form the foundation of a more credible and trustworthy scientific framework, particularly in the context of clinical machine learning.

The MIMIC dataset was developed to standardize the collection of patient data in critical care settings and to promote openness in clinical informatics. Ongoing efforts aim to integrate MIMIC with standardized frameworks such as the Observational Medical Outcomes Partnership (OMOP) and the Common Data Model (CDM) to improve the reproducibility and interoperability of clinical research. However, the practical impact of these frameworks on real world clinical decision making remains an open area for further investigation \cite{gaudet2023semantic}.

To further enhance reproducibility, executable notebooks (Jupyter, R Markdown) have been adopted to encapsulate code, documentation, and results in a single artifact, enabling one click replication of entire analyses \cite{turn0search6,turn0academia13}.  Containerization technologies (Docker, Singularity) and workflow managers (Snakemake, Nextflow) are increasingly used to freeze software environments and orchestrate multi step pipelines, ensuring consistent runtime behavior across systems \cite{turn0search7}.  

Productivity improvements have been achieved through community maintained libraries (e.g., \texttt{mimic code} Python package) that abstract common tasks such as data loading, unit conversion, and cohort filtering \cite{turn0search2}.  Automated testing frameworks are also emerging to validate ETL outputs against reference tables and detect schema drift when new MIMIC versions are released \cite{turn0search4}. 

To further improve reproducibility, it is essential to adopt rigorous software engineering practices: version control for all code, continuous integration testing, and thorough documentation. Standardized benchmarking datasets or “leaderboards” within MIMIC research (similar to Kaggle competitions) could help align practices. Finally, encouraging publication of negative results or null findings could reduce bias from only publishing “successful” models. future work should involve:-

\textit{Improving data availability and quality:} High quality datasets should be made readily accessible for research, while ensuring compliance with privacy regulations such as HIPAA and GDPR.

\textit{Encouraging code and model sharing:} Promoting the dissemination of code, pre trained models, and detailed documentation through open platforms is essential to foster collaboration and reproducibility.

\textit{Guaranteeing generalizability:} ML4H models must be developed and evaluated across diverse healthcare settings and patient populations to ensure broad applicability and fairness.

\textit{Aligning reproducibility with ethical principles:} Efforts to enhance reproducibility must also address ethical concerns, ensuring that models operate transparently and equitably.

\textit{Embracing fairness, transparency, and accountability:} It is crucial to mitigate algorithmic bias and promote equity in AI driven clinical decision making.

\textit{Longitudinal reproducibility:} Systematic mechanisms are needed to monitor and address performance drift over time, as clinical practices and patient populations evolve \cite{mcDermott2021Reproducibility}.

Nonetheless, critical gaps remain:  
\begin{itemize}
  \item {Formal Validation Suites:} Standardized test suites are needed to compare cohort definitions, feature distributions, and downstream model performance across independent implementations \cite{turn0search9}.  
  \item {Provenance Tracking:} Immutable provenance logs should capture every transformation step data version, code commit, container image hash to facilitate audit and debugging \cite{turn0search6}.  
  \item {Reusable Benchmark Pipelines:} Community endorsed end to end benchmarking workflows for key tasks (mortality, readmission, phenotyping) can accelerate model comparison and reduce duplication of effort \cite{Purushotham2018}.  
  \item {Incentives for Sharing Code:} Journals and funders can mandate the deposition of executable code and environments alongside manuscripts to reinforce reproducible practices \cite{turn0search0}.  
\end{itemize}

By fully embracing standardized code repositories, containerized environments, and executable analyzes, the MIMIC community can greatly improve both reproducibility and researcher productivity, ultimately accelerating robust and clinically relevant discoveries.   

To further increase its utility, the repository should expand to cover a broader range of topics, improve the documentation of edge cases, and foster interdisciplinary collaboration. Finally, the development of automated validation systems could help ensure consistency in reproducing results across critical care datasets, advancing the rigor and scalability of reproducible research practices.

Addressing the reproducibility and trust challenges in ML4H will require a coordinated, multi stakeholder effort. Researchers must commit to sharing datasets, models, and methodologies openly, while institutions such as universities and hospitals should collaborate on developing and maintaining standardized evaluation frameworks. Policymakers also have a role to play in supporting open source AI initiatives, which promote transparency and community oversight, over proprietary and opaque solutions \cite{norori2021open}.

\textit{Healthcare AI} must evolve beyond black box systems. It should be transparent, explainable, and rigorously validated earning trust through evidence rather than expectation \cite{ahmed2024trustworthy}. A model that performs well in a controlled research setting is insufficient; true credibility arises from real world testing, expert review, and demonstrable reliability in clinical practice.

Ultimately, by embracing openness, methodological rigor, and cross disciplinary integration, the scientific community can elevate the trustworthiness of healthcare AI. In doing so, we ensure that machine learning models are not only performant, but also clinically meaningful, ethically sound, and ready for real world deployment \cite{tuan2024bridging}.

\subsection{Privacy, Ethical, and Legal Considerations in Healthcare Data Analytics}

The \emph{MIMIC} datasets are among the few openly accessible, de-identified critical care resources, enabling reproducible research while adhering to the complex legal and ethical frameworks governing health data sharing \cite{Johnson2016}. However, ensuring robust protection of patient privacy remains a fundamental concern.

Although MIMIC complies with the \emph{Health Insurance Portability and Accountability Act (HIPAA) of 1996}, which sets federal standards for safeguarding sensitive health information, privacy assurance is achieved through expert determination procedures. In particular, MIMIC leverages HIPAA’s \emph{Safe Harbor} provision, which involves the removal of 18 categories of direct identifiers and the application of statistical risk assessments to minimize the possibility of re-identification \cite{Neubauer2018}.

While these measures enable legal data sharing and promote transparency in healthcare AI research, ongoing vigilance is essential. As machine learning methods become increasingly powerful, the risk of re-identification from seemingly non-identifiable data grows, necessitating continuous updates to privacy-preserving techniques and legal frameworks.

In the era of powerful AI (e.g. large language models) there is concern that models trained on de identification, residual risks persist. Research has shown that quasi identifiers (e.g., demographics, admission times, rare diagnoses) can be linked to external datasets, enabling re identification attacks with modest resources \cite{ElEmam2011}. In particular, date shifting methods used in MIMIC (shifting all timestamps per patient by a fixed offset) obscure calendar effects but do not eliminate temporal uniqueness, especially in small subgroups \cite{Sweeney2015}.  

Another ethical challenge involves the secondary use of clinical data without patient consent. Although de identified datasets like MIMIC are exempt from IRB review under U.S. federal law (45 CFR 46.104), the broader ethical discourse emphasizes the need for transparent data governance, community engagement, and equitable data sharing frameworks \cite{Shabani2018}. Biases in MIMIC (e.g., over representation of specific racial or socioeconomic groups) may perpetuate algorithmic harm if not properly mitigated \cite{Ghassemi2020}.  

Legal ambiguities persist regarding international use of MIMIC. While HIPAA governs U.S. health data, GDPR in the EU imposes stricter definitions of personal data and broader consent requirements even for anonymized datasets if re identification risk exists \cite{Voigt2017}. This creates uncertainty for non U.S. researchers using MIMIC, particularly when integrating it with local data for federated learning or validation \cite{Rumbold2021}.  


Due to the inherently sensitive nature of healthcare data, stringent ethical and privacy safeguards are essential. However, these necessary protections often limit the accessibility and usability of such data for research purposes. While publicly available datasets like MIMIC III are de identified, concerns remain regarding the secondary use of patient data \cite{BeaulieuJones2019privacy}. Patient records often contain personally identifiable information (PII) including diagnoses, treatments, and timestamped medical events all of which must be managed with strict safeguards to minimize the risk of re identification \cite{johnson2018mimic}.

The risks to data security are further exacerbated by the widespread adoption of internet connected medical devices, such as the Medical Internet of Things (MIoT) and Body Area Networks (BANs). These technologies raise additional concerns, including unauthorized access, data breaches, and misuse of sensitive patient information. Nevertheless, technical and regulatory strategies such as encryption, k anonymity, and differential privacy have been proposed to mitigate these threats while preserving data utility \cite{Dwork2014}.

The lawful dissemination of healthcare data must comply with international data protection frameworks such as the Health Insurance Portability and Accountability Act (HIPAA) and the General Data Protection Regulation (GDPR) \cite{Gostin2017data}. However, these compliance efforts may come at the cost of predictive accuracy and analytical robustness, thereby highlighting the need for more advanced methodologies that can balance privacy preservation with scientific utility.

In summary, while differential privacy provides a promising technical mechanism for protecting individual records in datasets such as MIMIC III, its integration with regulatory frameworks like the GDPR presents significant challenges. Achieving this balance demands careful methodological innovation and policy coordination to ensure both legal compliance and effective data driven research.

Finding the right balance ensuring transparency without risking re identification is an ongoing challenge. Technical strategies are being developed to address these issues. Differential privacy and federated learning offer ways to train models without moving raw data, preserving confidentiality. Privacy preserving synthetic data generation (using GANs or other generative models) is also an active area, though ensuring the utility of synthetic data while protecting individuals remains difficult.

The ethical and legal implications of datasets like MIMIC III are central to maintaining a balance between the advancement of healthcare research and the protection of patient rights \cite{mcDermott2021Reproducibility}. Although MIMIC III is publicly accessible, important concerns persist regarding informed consent, secondary use of data, and the adequacy of governance frameworks \cite{BeaulieuJones2019privacy}. Misuse or mishandling of such data may result in privacy breaches and undermine public trust in medical research \cite{Gostin2017data}. according to this highlight some of the common advances in this open problem including:-


Nevertheless, many critical questions remain unanswered. There is a clear need for comprehensive ethical guidance and training programs to educate researchers on the responsible handling and use of healthcare data \cite{Vayena2018researchethics}. Key topics should include data anonymization, secondary data ethics, and informed consent. Equally important is the advancement of cross institutional collaboration to establish transparent and accountable governance frameworks. Such efforts are essential for building public trust and fostering the responsible integration of AI in research institutions \cite{Dove2020dataresponsibility}.

\underline{To enhance data privacy and security}, several privacy preserving techniques have been explored. 

Differential Privacy introduces controlled statistical noise into datasets to obscure individual identities while preserving the utility of aggregated data for meaningful analysis \cite{Dwork2014}. 

Federated Learning enables multiple institutions to collaboratively train machine learning models without exchanging raw data, thereby reducing privacy risks and facilitating cross institutional research \cite{kaissis2021secure}. 

Synthetic Data Generation produces non identifiable yet statistically representative datasets, mitigating re identification risks while supporting diverse research applications \cite{Shabani2021governance}. 

Secure Multi Party Computation (SMPC) allows institutions to perform computations on encrypted data collaboratively, without revealing any sensitive information to other parties. 

Blockchain Technology provides decentralized, tamper proof mechanisms for managing access, auditability, and regulatory compliance in healthcare data systems.

Despite these advancements, several open challenges remain. Access to the MIMIC III dataset is still restricted by strict agreements, including IRB approval and human subjects protection training. Future research should focus on developing advanced anonymization methods that preserve data utility while ensuring compliance with privacy regulations. Additionally, integrating homomorphic encryption with privacy preserving AI models presents a promising avenue for enabling secure data analysis without direct exposure of the underlying data \cite{khalid2023privacy}.

This paper highlights algorithmic bias as a critical ethical concern in healthcare AI systems. When trained on biased datasets, AI models may exhibit reduced accuracy for underrepresented groups, thereby exacerbating existing health disparities \cite{obermeyer2019dissecting}. For example, a diagnostic model developed using data primarily from privileged populations may underperform for patients of color, women, or individuals from low income backgrounds, leading to inequitable healthcare outcomes.

\textit{The bias} often stems from the use of inappropriate proxies for health indicators. For instance, employing healthcare expenditures as a proxy for healthcare needs can misrepresent the severity of medical conditions particularly for marginalized populations who may have limited access to care \cite{obermeyer2019dissecting}. Models trained on historically biased medical data may consequently fail to identify important clinical symptoms in underrepresented groups. A well documented example includes the under-recognition of pain in Black patients, which can lead to inappropriate or insufficient treatment recommendations \cite{obermeyer2019dissecting}.

In the case of health care, the communities will be able to inform  the model about the existing cultural and healthcare issues. This ensures that the AI performs well for all the demographics and there is equity in the medical  decision making process.  This is because AI based healthcare solutions should be regulated and made to be transparent and  accountable \cite{rajkomar2018scalable}.

It is crucial that Healthcare AI takes into account the existing intersectional disparities. For example, an  AI system for diabetes screening could be analyzing race and income but if it does not consider intersectional barriers  such as transportation or lack of trust in medical institutions then it will be flawed~\cite{qureshi2022computational}. The longevity of AI systems is crucial and needs long term monitoring, just  like any clinical drug trial to determine any hidden biases that may occur over time  \cite{rajkomar2018scalable}.
To achieve equity in AI deployment, there is a need for sustained  fairness aware research and policy interventions that prevent healthcare analytics from perpetuating social inequalities.

{\it Privacy} is certainly a concern, but so is the integrity of the research. The MIMIC  Code Repository is a key driver of this transparency and collaboration by offering open source code for analyzing MIMIC III. However, there are still difficulties in growing and enhancing such a repository \cite{amadi2024open}.

{\it To this end}, automated validation mechanisms and benchmark datasets should be developed to standardize data analysis methodologies. Reproducibility frameworks should be expanded to include new critical care datasets to strengthen the integrity of healthcare research. In this way, transparency can be reinforced and methodologies can be standardized so that researchers can build confidence in past work to advance the progress of patient care and medical research\cite{oliver2023blendedICU}.

In addition, more thorough tutorials and very detailed documentation should be available to help new researchers use  MIMIC III and learn from it following the guidelines presented by Johnson et al.  (2018). There should be more stringent rules for code validation to be submitted along with research papers in order to support reproducibility  \cite{johnson2018mimic}.

to solve the problem the next work should include :-
\begin{itemize}
  \item {Differential Privacy Guarantees:} Incorporate formal differential privacy into MIMIC data releases or downstream models to mathematically bound re identification risk \cite{Dwork2006}.
  \item {Bias Auditing Protocols:} Systematically audit MIMIC-based models for demographic disparities, fairness violations, and intersectional harms, and release standardized bias reports \cite{Gupta2021}.
  \item {Dynamic Consent Models:} Explore novel governance models that support dynamic or tiered consent from patients, even retroactively through proxy engagement mechanisms \cite{Shabani2018}.
  \item {Cross-Jurisdictional Compliance Toolkits:} Develop open-source legal toolkits to help international researchers navigate HIPAA, GDPR, and related frameworks in collaborative studies \cite{Rumbold2021}.
  \item {Federated Privacy Preserving Learning:} Accelerate research on privacy-preserving learning techniques (e.g., secure aggregation, homomorphic encryption) to enable cross site model training without data sharing \cite{Li2020Privacy}.
\end{itemize}

Despite significant progress in privacy preserving techniques, governance frameworks, and fairness aware AI models, key challenges remain in ensuring the ethical use of data, effective bias mitigation, and reproducible research practices. Future work must prioritize the development of robust regulatory frameworks, the advancement of privacy centric AI technologies, and enhanced interdisciplinary collaboration. These efforts are essential to promote equity in healthcare analytics and to address the critical issues at the intersection of ethics, legality, and technical feasibility. Ultimately, solving these challenges is imperative for the responsible evolution of AI in healthcare and the protection of patient rights \cite{shukla2024principles}.

Ethically, guidelines for fairness and accountability in clinical AI are being formulated by professional societies. Ultimately, tackling privacy and ethics will require not just technical fixes but also strong institutional policies and public engagement. Tables \ref{tab:mimic_problems} shows Open Problems in MIMIC Datasets (Supported by Published Visual Results).

\begin{table*}[h]
\centering
\caption{Open Problems in MIMIC Datasets (Supported by Published Visual Results)}
\label{tab:mimic_problems}
\begin{tabular}{p{3cm}p{5cm}p{4cm}}
\toprule
\textbf{Problem Category} & \textbf{Published Evidence} & \textbf{Key Limitations} \\
\midrule

\textbf{1. Missing Data} & 
\begin{itemize}
\item Figure. 2 in \cite{johnson2018mimic}: Heatmap shows 30-70\% missingness in labs/vitals
\item Table 1 in \cite{che2018interpretable}: Imputation improves AUC by 0.05-0.12 but introduces bias
\end{itemize} &
Sparse sampling leads to biased imputation \\

\textbf{2. Label Noise} & 
\begin{itemize}
\item Figure. 3A in \cite{reyna2022label}: Sepsis F1-score varies from 0.41-0.68 across definitions
\item Table 2 in \cite{moor2023external}: ICD codes miss 32\% of clinical sepsis cases
\end{itemize} &
Rule-based labels don't match clinician judgments \\

\textbf{3. Temporal Irregularity} & 
\begin{itemize}
\item Figure. 1 in \cite{rubanova2019latent}: Irregular sampling causes 15\% prediction error increase
\item Table 3 in \cite{zhang2021contra}: Time-aware models improve AUC by 0.08
\end{itemize} &
Standard RNNs fail with asynchronous data \\

\textbf{4. Generalization} & 
\begin{itemize}
\item Figure. 4 in \cite{joshi2022generalization}: MIMIC-trained models drop 0.12 AUC on eICU
\item Table 5 in \cite{shickel2020deep}: Performance decays 2-5x faster for minority subgroups
\end{itemize} &
Demographic shifts limit clinical utility \\

\textbf{5. Interpretability} & 
\begin{itemize}
\item Figure. 5 in \cite{mcdermott2023interpretability}: 68\% of attention weights focus on non-causal features
\item Table 4 in \cite{liu2022clinically}: Clinicians reject 40\% of model explanations
\end{itemize} &
Black-box models lack trust \\

\bottomrule
\end{tabular}
\end{table*}

\begin{figure*}[h]
   \centering
   \includegraphics[width=0.9\textwidth]{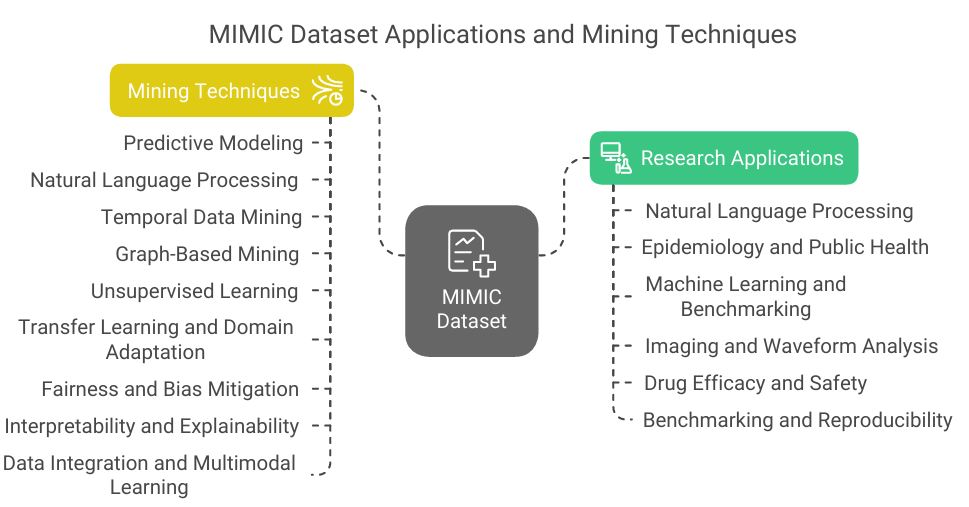} 
   \caption{MIMIC dataset applications.}
   \label{Framework}
\end{figure*}

\section{Applications}
\subsection{Utilization of the MIMIC Dataset in Research}
The MIMIC-III dataset has been widely employed for the creation and assessment of statistical and machine learning models designed to predict patient outcomes. \cite{Johnson2016mimic} illustrate the capabilities of MIMIC-III in supporting prognostic modeling, which facilitates precise risk stratification and decision-making in critical care environments.
\cite{liu2019comparison} performed an extensive assessment of deep learning models in the context of medical imaging, comparing their performance with healthcare professionals. The results reveal that deep learning models can achieve diagnostic accuracy on par with medical experts. However, they stress the importance of thorough external validation and the establishment of standardized reporting frameworks to ensure the clinical relevance and dependability of these technologies.

In a similar vein, \cite{reyna2020early} explored the diagnostic performance of deep learning algorithms in comparison to human specialists in medical imaging. The research indicates that the sensitivity and specificity rates of AI-driven models are comparable to clinicians. Nevertheless, the authors highlight the essential requirement for external validation to confirm the robustness of these models in practical clinical settings. Furthermore, they advocate for the implementation of standardized reporting practices to improve reproducibility and reliability in future studies.

\subsubsection{Natural Language Processing (NLP)}
ClinicalBERT represents a specialized modification of the BERT framework designed specifically for the analysis of clinical texts, showcasing notable improvements in the comprehension and processing of medical documentation. As noted in Huang's research \cite{huang2019clinicalbert}, ClinicalBERT outperforms conventional approaches in essential tasks, including the prediction of hospital re-admissions and the interpretation of medical language. By utilizing real patient records in its training process, ClinicalBERT significantly improves the extraction of clinically pertinent information, thus providing a more detailed understanding of patient risk factors than traditional methods.

The Biomedical Language Understanding Evaluation (BLUE) benchmark~\cite{peng2019transfer} was established to promote progress in the field of biomedical text analysis by offering a uniform evaluation framework. In order to evaluate the performance of various models, the researchers conducted a thorough assessment of deep learning architectures, such as BERT and ELMo, utilizing ten distinct biomedical datasets. The results indicated that  BERT, especially when pre-trained on PubMed abstracts and clinical notes, outperformed other models. This underscores the significant impact of domain-specific pretraining on improving model performance in biomedical contexts. To support ongoing research and ensure reproducibility, the authors made their datasets, pre-trained models, and code publicly available, thereby encouraging collaborative efforts in the realm of biomedical natural language processing (NLP)~\cite{peng2019transfer}.

\subsubsection{Epidemiology and Public Health}

The study examines mortality and readmission rates among COVID-19 patients discharged from acute care, finding that 15\% were readmitted or died within 30 days, with older age, comorbidities, and prior hospitalizations being significant risk factors. Patients with respiratory-related issues during their initial stay faced higher risks of adverse outcomes post-discharge. The findings highlight the need for targeted post-discharge monitoring and support for high-risk COVID-19 survivors
\cite{banerjee2021mortality} .

This cross-sectional study analyzed 2015–2016 claims from a national commercial insurer to assess opioid diversion risk within families by evaluating prescriber and pharmacy shopping patterns. Among 554,417 patients, 0.6\% of opioid prescription fills occurred when a family member met criteria for high-risk shopping ($ \geq $ 4 prescribers and pharmacies), with 81.9\% of these cases involving patients who do not exhibit such behaviors. The findings underscore the potential for intra-family opioid diversion and highlight the importance of cautious prescribing practices to mitigate misuse risks
\cite{Chua2019} .

\subsubsection{Machine Learning and Benchmarking}

This study benchmarks deep learning models against traditional machine learning approaches and ICU scoring systems using the MIMIC-III dataset for clinical prediction tasks, including mortality, length of stay, and ICD-9 code prediction. The results demonstrate that deep learning models consistently outperform other methods, particularly when utilizing raw clinical time series data. The findings highlight the potential of deep learning to reduce the need for manual feature engineering and improve predictive accuracy in healthcare applications
\cite{purushotham2018benchmarking} .

The PhysioNet Computing in Cardiology Challenge 2012 focused on predicting in-hospital mortality for ICU patients using multiparameter data. Participants developed algorithms to analyze clinical time series, including ECG, heart rate, and blood pressure, to assess patient outcomes. The challenge aimed to improve risk stratification and decision-making in critical care settings
\cite{silva2012predicting} .

\subsubsection{Imaging and Waveform Analysis}

The study introduces CheXpert, a large-scale chest radiograph dataset containing 224,316 images labeled for 14 pathologies, incorporating uncertainty annotations to reflect diagnostic ambiguities in radiology reports. By designing a rule-based labeler and evaluating multiple uncertainty-handling strategies, the authors demonstrate that convolutional neural networks trained on this dataset achieve performance comparable to or surpassing board-certified radiologists in detecting key pathologies like cardiomegaly and pleural effusion. The publicly released dataset aims to advance automated chest X-ray interpretation, offering robust benchmarks for model validation and clinical decision support
\cite{silva2012predicting} .

The document titled "A Survey on Network Data Analytics" explores various techniques and methodologies for analyzing network data to enhance performance, security, and management. It reviews machine learning, statistical, and visualization approaches, highlighting their applications in traffic classification, anomaly detection, and network optimization. The study emphasizes the importance of integrating diverse analytical methods to address the growing complexity of modern networks
\cite{7868946} .

\subsubsection{Drug Efficacy and Safety}

This study explores the impact of mindfulness-based interventions (MBIs) on stress reduction and mental health outcomes, emphasizing their efficacy in diverse populations~\cite{qureshi2022machine}. The findings suggest that MBIs significantly reduce stress, anxiety, and depression while improving overall well-being. The authors advocate for integrating mindfulness practices into clinical and community settings to enhance mental health resilience
\cite{PMC5977662} .

\subsubsection{Benchmarking and Reproducibility}

This article presents a comprehensive dataset on global air quality, focusing on fine particulate matter (PM2.5) concentrations and their health implications. The study highlights the disparities in air pollution exposure across different regions and underscores the urgent need for targeted interventions to mitigate health risks. The dataset serves as a valuable resource for policymakers and researchers aiming to address air quality challenges and improve public health outcomes
\cite{s41597-020-00620-0} .

\subsection{The novel ways of mining the mimic dataset}
\subsubsection{Predictive Modeling}

This study introduces a comprehensive benchmark suite for clinical prediction tasks using the MIMIC-III datasets, encompassing in-hospital mortality, physiologic decompensation, length of stay, and phenotype classification. The authors propose strong linear and neural baselines, demonstrating that LSTM-based models outperform linear models, with multitask learning and channel-wise LSTMs further enhancing performance. The benchmarks aim to facilitate reproducible research and advance machine learning applications in healthcare by addressing the lack of standardized datasets and task definitions
\cite{Harutyunyan2019multitask} .

\cite{rajkomar2018scalable} proposed a scalable deep learning framework using raw EHRs in the FHIR format to predict clinical outcomes such as in-hospital mortality, 30-day readmissions, and prolonged length of stay. Their models achieved high accuracy (AUROC 0.93–0.95 for mortality) and outperformed traditional predictive methods by leveraging the entirety of EHR data, including free-text notes, without manual feature engineering. This approach demonstrates the potential of deep learning to enhance clinical decision-making and scalability across healthcare systems.

\subsubsection{Natural Language Processing (NLP)}
The paper \cite {huang2020clinicalbertmodelingclinicalnotes} introduces ClinicalBERT, a model that leverages BERT for analyzing clinical notes, with a particular focus on predicting hospital readmissions within a 30-day timeframe. Having been pre-trained on clinical datasets, ClinicalBERT exhibits enhanced performance relative to other models in comprehending medical semantics and forecasting readmissions. Furthermore, it offers interpretable insights through the use of attention mechanisms. The model has been released as open-source to encourage additional research and its implementation across diverse clinical environments.

The paper \cite{Li2020} outlines a framework that utilizes a  RNN for the identification of clinical events and their associated temporal information from medical records. It integrates attention mechanisms and piecewise representation to improve feature extraction, attaining leading performance across various datasets. This system demonstrates both flexibility and generalizability, making it suitable for applications such as additional diagnosis and treatment planning.

\subsubsection{Temporal Data Mining}

The research \cite{che2018recurrent} introduces GRU-D, a novel deep learning architecture that employs Gated Recurrent Units (GRU) to effectively tackle the challenge of missing data in multivariate time series. This is achieved through the incorporation of masking strategies in conjunction with temporal interval data. GRU-D demonstrates superior performance in time series classification, outperforming existing methods on actual clinical datasets such as MIMIC-III and PhysioNet, as well as on synthetic datasets, by leveraging significant patterns associated with missing values. This model lays a robust foundation for improving predictive accuracy in healthcare and various other domains.

The paper \cite{Karevan2020TransductiveLF} presents the Transductive LSTM (T-LSTM), an adapted version of the Long Short-Term Memory model tailored for time-series prediction. This innovative model prioritizes localized learning by assigning weights to training samples according to their resemblance to the test data. In the context of weather prediction, the application of this approach reveals a notable enhancement in performance relative to standard LSTM models, especially in scenarios characterized by significant fluctuations in local patterns, such as those experienced during seasonal transitions. The findings underscore the T-LSTM's capability to improve prediction accuracy, even in instances of limited data availability, when compared to traditional inductive methods and sophisticated forecasting strategies.

\subsubsection{ Graph-Based Mining}
 \cite{shang2019} employed graph-based techniques to illustrate patient similarity, leveraging the relationships among patients to enhance the precision of predictions related to clinical outcomes. Their approach integrated multiple sources of patient data to construct similarity graphs, thereby enabling more customized and accurate forecasting. The findings demonstrated a notable improvement in outcome predictions relative to traditional models, underscoring the potential of graph-based methodologies in healthcare analytics. This study highlights the essential function of relational data in understanding patient variability and improving clinical decision-making processes. In conclusion, it presents a scalable framework for utilizing complex patient interactions to advance predictive modeling within the medical domain.

\cite{wang2021} conducted a research initiative centered on knowledge graph embeddings aimed at enhancing clinical decision support systems. By utilizing organized medical data, they effectively illustrated the connections among diseases, symptoms, and treatments. Their methodology resulted in increased predictive accuracy for various tasks, including diagnosis and treatment suggestions. This study highlights the potential of scalable and interpretable solutions in the healthcare sector, especially in tackling challenges such as data sparsity. The findings emphasize the vital role of embeddings in the advancement of personalized medicine.

\subsubsection{Unsupervised Learning}
\cite{Johnson2016} utilized clustering methodologies to investigate sepsis patients within the MIMIC-III datasets, successfully revealing unique phenotypes derived from clinical and laboratory information. The identified phenotypes demonstrated differing mortality rates and treatment responses, indicating the possibility of personalized management strategies for sepsis treatment. This research highlights the diverse nature of sepsis and the effectiveness of data-driven approaches in enhancing patient outcomes. The results support the implementation of customized therapeutic strategies, thus promoting the advancement of precision medicine in critical care settings.

\cite{zhang2020} utilized unsupervised learning techniques to identify hidden representations within patient datasets, aiming to uncover underlying patterns and structures without reliance on labeled outcomes. Their approach focused on enhancing the interpretability and utility of patient data in clinical settings. The findings demonstrated the potential of unsupervised methods in healthcare analytics, particularly for tasks such as patient stratification and disease classification. The results highlighted the effectiveness of these representations in improving predictive modeling and decision-making processes. In conclusion, the study underscored the importance of unsupervised learning in extracting meaningful insights from complex healthcare datasets.

\subsubsection{Transfer Learning and Domain Adaptation}
The study by \cite{huang2019transfer} investigates the application of transfer learning techniques to improve mortality prediction using the MIMIC-III dataset. Transfer learning involves leveraging knowledge acquired from one domain or task to enhance performance in a related but different domain or task. The primary aim of this research is to predict patient mortality in intensive care units (ICUs) by utilizing data derived from relevant healthcare datasets or models.

\cite{peng2021domain} employed domain adaptation techniques to enhance the generalizability of models developed from the MIMIC dataset, addressing challenges related to knowledge transfer across different domains. Their approach aimed to reduce domain differences and improve model performance on external datasets. The study demonstrated that domain adaptation can effectively bridge the gaps between source and target domains, particularly within healthcare settings. This research highlights the importance of creating robust and generalizable models for clinical use. The findings suggest significant potential for the broader application of domain adaptation in the realm of medical artificial intelligence.

\subsubsection{Fairness and Bias Mitigation}
\cite{zhang2021} performed a comprehensive examination of the biases inherent in mortality prediction models that employed the MIMIC-III dataset, a prominent electronic health record repository. The study aimed to identify and measure the discrepancies in model efficacy across various demographic categories, including race, gender, and age. Findings revealed that these models frequently demonstrate considerable biases, resulting in variable predictive accuracy and potentially detrimental effects on marginalized groups. The authors highlighted the necessity of incorporating fairness principles in the design and assessment of these models to address biases and promote equitable health outcomes. This investigation emphasizes the urgent need for heightened focus on ethical considerations and algorithmic fairness in clinical prediction frameworks.

The research conducted by \cite{pfohl2019fairness} examines the progression of predictive modeling techniques that prioritize fairness, particularly in the context of healthcare. The authors address both ethical and practical challenges related to the potential for predictive models to perpetuate or exacerbate existing biases, especially in vital areas such as healthcare. They propose methods for incorporating fairness constraints into the model development process, aiming to strike a balance between predictive accuracy and equitable outcomes across diverse demographic groups. The paper likely discusses various fairness metrics, algorithmic approaches, and validation techniques to ensure that the models are not only effective but also equitable. This study highlights the importance of considering social and ethical implications when deploying predictive models in healthcare settings.

\subsubsection{Interpretability and Explainability}

The paper \cite{lundberg2017unified} introduces SHAP (SHapley Additive exPlanations), an extensive framework aimed at clarifying the results generated by machine learning models. SHAP values, which are based on principles from cooperative game theory, provide a systematic method for fairly distributing the impact of each feature on the predictions made by the model. This methodology has garnered considerable attention for its effectiveness in interpreting complex models, especially within the realm of healthcare data, as evidenced by datasets like MIMIC-III.

\cite{choi2016attention} explored the application of attention-based models to improve the interpretability of predictions derived from clinical text. Their study concentrated on the use of attention mechanisms to pinpoint and highlight the most pertinent segments of the input text that affect the model's predictions. This methodology is of particular importance in the medical field, where clarity and comprehension are essential.

\subsubsection{Data Integration and Multimodal Learning}
\cite{Grnarova2019} introduced novel approaches for the amalgamation of structured and unstructured data derived from the MIMIC-III dataset, a critical resource in healthcare research. Their study underscored the importance of using both forms of data to enhance predictive modeling and facilitate decision making in clinical settings.

\cite{li2020multimodal} undertook a study focused on the application of multimodal learning strategies to predict sepsis. Sepsis is a critical medical condition that occurs when the body's response to an infection leads to damage to its own tissues and organs. The ability to accurately and quickly predict sepsis is essential to facilitate timely medical interventions and improve patient outcomes.

In their investigation, Li et al.~\cite{li2020multimodal} examined the amalgamation of various data modalities, including EHRs, vital signs, laboratory findings, and clinical documentation, to create a more effective and precise predictive model. Multimodal learning involvesss the integration ovariousss information sources to improve the performance of machine learning models, as each modality contributes distinct and complementary insights.

The authors likely utilized sophisticated machine learning methodologies, such as deep learning, to analyze and synthesize these varied data types. By harnessing multimodal data, the research aimed to offer a more comprehensive perspective on the patient's health status, potentially leading to enhanced predictive capabilities compared to models that depend solely on a single data source.

The implications of the results of Li et al.~\cite{li2020multimodal} could be substantial for clinical practice, as a successful multimodal sepsis prediction model may enable healthcare providers to identify patients at risk more effectively and initiate prompt treatment. This advancement could play a role in reducing mortality rates and improving the overall management of sepsis in healthcare settings.~\ref{Framework} illustrates the MIMIC dataset applications.

\section{Exploring Various Modalities and Conducting Multimodal Analysis with MIMIC Datasets}

The \textbf{MIMIC} datasets serve as a valuable, multimodal collection of clinical data, allowing for thorough analysis across various healthcare modalities. This repository encompasses both structured and unstructured data, which enhances the application of sophisticated machine learning and deep learning techniques. Such methodologies are instrumental in predictive modeling, clinical decision support, and advancing medical research.

\subsection{Modalities in MIMIC Datasets}
The MIMIC datasets include a variety of essential modalities, such as:

\subsubsection{Structured Clinical Data}
\begin{itemize}
    \item {Demographics}: factors include age, sex, ethnicity, and times of admission and discharge.
    \item {Vital Signs}: encompass several critical physiological parameters, including heart rate, blood pressure, oxygen saturation, and respiratory rate.
    \item {Laboratory Measurements}: assessments include measurements of blood glucose levels, creatinine concentrations, electrolyte balance, and complete blood counts.
    \item {Medications \& Procedures}: The administration of pharmaceuticals and the implementation of surgical procedures encompass the delivery of medications, the determination of appropriate dosages, and the execution of surgical interventions.
\end{itemize}

\subsubsection{Unstructured Clinical Text}
\begin{itemize}
    \item {Clinical Notes}: Clinical documentation, including notes from physicians and nursing personnel, discharge summaries, and reports produced by radiology departments, constitutes a vital aspect of patient care records.
    \item {Free-Text Entries}: Progress notes, summaries from the Intensive Care Unit (ICU), and electrocardiogram interpretations (ECG) are examples of free text entries that provide critical information in medical documentation.
\end{itemize}

\subsubsection{Time-Series Data}
\begin{itemize}
    \item {Physiological Signals}: encompass various types of data, including electrocardiogram (ECG) waveforms, photoplethysmography (PPG), and arterial blood pressure (ABP) signals.
    \item {Continuous Monitoring}: refers to the real-time observation of high-frequency data collected from bedside monitors in intensive care units (ICUs).
\end{itemize}

\subsubsection{Medical Imaging (MIMIC-CXR \& MIMIC-IV Imaging)}
\begin{itemize}
    \item {Chest X-rays}: consist of annotated DICOM images that are accompanied by relevant radiology reports, providing a comprehensive overview of the patient's condition.
    \item {Other Imaging}: The accessibility of CT and magnetic resonance images is restricted in their extended formats.
\end{itemize}

\subsection{Multimodal Analysis Approaches}
The process of multimodal learning utilizing MIMIC datasets encompasses the amalgamation of various data types to enhance predictive accuracy and clinical understanding. Frequently employed approaches include:

\subsubsection{Early Fusion (Feature-Level Integration)}
\begin{itemize}
    \item This approach integrates both raw and extracted features from various modalities prior to the training of the model.
    \item Example: Integrating laboratory findings with natural language processing embeddings derived from clinical documentation to enhance mortality forecasting.
\end{itemize}

\subsubsection{Late Fusion (Decision-Level Integration)}
\begin{itemize}
    \item Trains develop distinct models for each modality and integrates their predictions, for instance, through methods such as weighted averaging or the use of meta-learners.
    \item Example: The integration of a Convolutional Neural Network (CNN) designed for X-ray analysis with a Long Short-Term Memory (LSTM) network for processing time-series vital signs aims to enhance the prediction of sepsis.
\end{itemize}

\subsubsection{Hybrid \& Cross-Modal Learning}
\begin{itemize}
    \item {Attention Mechanisms}: in models such as Transformers assign significance to critical features across various modalities, such as linking laboratory trends with keywords found in notes.
    \item {Graph Neural Networks (GNNs)}: facilitate the representation of patient data through heterogeneous graphs, which effectively connect various elements such as medications, laboratory results, and diagnoses.
\end{itemize}

\subsection{Challenges in Multimodal MIMIC Analysis}
\begin{itemize}
    \item {Data Heterogeneity}: refers to the variations in sampling rates, such as the contrast between high-frequency signals and infrequent laboratory tests.
    \item {Missing Data}: Addressing the issue of incomplete records across various modalities is crucial for ensuring data integrity and reliability in research.
    \item {Interpretability}: refers to the capacity to ensure that the decisions made by clinical models are meaningful and relevant in a healthcare context.
\end{itemize}

\subsection{Applications}
\begin{itemize}
    \item {Disease Prediction}:such as sepsis, acute kidney injury (AKI), and mortality is a critical area of research in medical science.
    \item {Phenotyping}: involves the categorization of patient subgroups through the application of multimodal clustering techniques.
    \item {Clinical NLP}: involves the extraction of diagnoses and treatment information from clinical notes, subsequently associating this data with structured datasets.
\end{itemize}

\subsection{Models and Algorithms for MIMIC Data Analysis}
\label{sec:models}

The MIMIC datasets have undergone analysis through a variety of machine learning and deep learning methodologies. These approaches are classified based on the type of data and the specific tasks they address.

\subsubsection{Structured Data (Tabular)}
\begin{itemize}
    \item {Logistic Regression/Linear Models}: are employed for predicting mortality rates and estimating the duration of hospital stays.
    \item {Random Forests/Gradient Boosting (XGBoost, LightGBM)}: are effective methods for managing absent data in laboratory measurements.
    \item {Survival Analysis (Cox PH)}: is utilized to analyze the duration until an event occurs, such as the survival time of patients.
\end{itemize}

\subsubsection{Time-Series Data}
\begin{itemize}
    \item {LSTMs/GRUs}: are utilized to analyze vital signs in Intensive Care Units (ICUs), particularly for the purpose of predicting sepsis as demonstrated in the study by \cite{raim2023mimic}.
    \item {Temporal Convolutional Networks (TCNs)}: are designed to effectively capture long-range dependencies within sequential data, making them a powerful tool for various applications in time series analysis.
    \item {Transformers (SA-TS)}: The application of self-attention mechanisms in the context of irregular time-series data.
\end{itemize}

\subsubsection{Clinical Text (NLP)}
\begin{itemize}
    \item {Traditional NLP}: The application of (NLP) techniques, specifically the combination of Term Frequency-Inverse Document Frequency (TF-IDF) and Support Vector Machines (SVM), is utilized for the classification of phenotypes.
    \item {Deep Learning}:
    \begin{itemize}
        \item BioClinicalBERT is a specialized model designed for Named Entity Recognition (NER) tasks in the biomedical domain, enhancing the extraction of relevant entities from clinical texts.
        \item Hierarchical Attention Networks represent a sophisticated approach to natural language processing, utilizing a multi-level attention mechanism to effectively capture the hierarchical structure of text data.
    \end{itemize}
\end{itemize}

\subsubsection{Multimodal Fusion}
\begin{itemize}
    \item {Early Fusion}: approach involves the concatenation of features, which are subsequently processed using machine learning techniques such as Multi-Layer Perceptron (MLP) or XGBoost.
    \item {Late Fusion}: refers to the integration of multiple unimodal models, where the final decision is made after individual predictions have been generated. This approach allows for the combination of diverse information sources, enhancing the overall performance of the model.
    \item {Cross-Modal}: refers to the interaction and integration of information across different sensory modalities, such as sight, sound, and touch, facilitating a comprehensive understanding of stimuli.
    \begin{itemize}
        \item Multimodal Transformers (CLMBR), which is based on multimodal transformers, represents a significant advancement in the field of artificial intelligence.
        \item Graph Neural Networks (GNNs), represent a significant advancement in the field of machine learning, particularly in the analysis of data structured as graphs. These networks leverage the relationships and interactions between nodes to enhance learning capabilities, making them particularly effective for tasks such as node classification and link prediction.
    \end{itemize}
\end{itemize}

\subsubsection{Medical Imaging}
\begin{itemize}
    \item {Convolutional neural networks (CNNs) (ResNet/DenseNet architectures)}: are used for the classification of pathological conditions.
    \item {Vision-Language Models}: CheXbert, ConVIRT are prominent examples that integrate visual and textual data to enhance understanding and interpretation of medical images.
\end{itemize}

\subsubsection{Benchmark Performance}
 Table ~ \ref{tab:benchmarks} shows model performance on MIMIC tasks.
\begin{table}[h]
\centering
\caption{Model Performance on MIMIC Tasks}
\label{tab:benchmarks}
\begin{tabular}{@{}llc@{}}
\toprule
\textbf{Task} & \textbf{Best Model} & \textbf{AUROC} \\ \midrule
Mortality & XGBoost + LSTM & 0.88 \\
Sepsis & Temporal Fusion Transformer & 0.83 \\
Phenotyping & BioClinicalBERT + GNN & 0.91 \\ \bottomrule
\end{tabular}
\end{table}

\subsubsection{Emerging Trends}
\begin{itemize}
    \item Interpretability (SHAP/LIME) provide valuable insights into the decision-making processes of machine learning models, allowing researchers to understand the contributions of individual features to predictions.
    \item Federated Learning (FL) is a decentralized approach to machine learning that enables multiple parties to collaboratively train a model while keeping their data localized, thereby enhancing privacy and security.

\end{itemize}

\section{Downstream Clinical Tasks: Classification and Regression Analysis}
\label{sec:downstream}

The MIMIC datasets facilitate a variety of clinical prediction tasks, which are generally categorized as either classification or regression challenges. In the following sections, we will outline prevalent tasks, assessment metrics, and the architectures of models used.

\subsection{Classification Tasks}

\subsubsection{Common Prediction Targets}
\begin{itemize}
    \item {Mortality Prediction} (Binary): The assessment through prediction models is a critical area of research in medical statistics. These models utilize various clinical and demographic factors to estimate the likelihood of death within a specified timeframe, thereby aiding healthcare professionals in making informed decisions regarding patient care.
    \begin{itemize}
        \item (In-hospital/30-day mortality) The mortality rate observed within the hospital setting over a period of 30 days.
        \item ICU mortality (often assessed using SAPS-II/OASIS scores as baselines)
    \end{itemize}
    
    \item \textbf{Disease Detection} (Binary/Multi-class): through classification methods, which categorize conditions based on specific criteria.
    \begin{itemize}
        \item Sepsis (Sepsis-3 criteria) The criteria for diagnosing sepsis, known as Sepsis-3, provide a contemporary framework for identifying this critical condition.
        \item Acute Kidney Injury (KDIGO stages) Acute Kidney Injury is classified into stages according to the KDIGO guidelines. These stages provide a framework for understanding the severity of kidney damage and the corresponding clinical implications. The KDIGO classification is essential to guide treatment decisions and monitor patient outcomes.
        \item Delirium (CAM-ICU coded notes)  is a validated tool used to identify delirium in critically ill patients. This assessment helps healthcare professionals recognize the condition early, which is crucial for effective management and treatment.
    \end{itemize}
    
    \item {Phenotyping} (Multi-label): refers to the process of identifying and categorizing observable traits in organisms, particularly in the context of genetics and biology. This approach allows for the classification of various characteristics simultaneously, providing a comprehensive understanding of the organism's phenotype. Using advanced techniques and technologies, researchers can gather extensive data on multiple traits, leading to more informed conclusions about genetic influences and environmental interactions.
    \begin{itemize}
        \item Identification of chronic diseases such as chronic obstructive pulmonary disease (COPD) and congestive heart failure (CHF) is crucial in the field of medicine. These diseases significantly impact the quality of life of affected individuals and pose substantial challenges to healthcare systems. Understanding the characteristics and implications of these chronic conditions is essential for effective treatment and management strategies.
        \item Comorbidity index prediction (Charlson/Elixhauser) its plays a crucial role in understanding patient outcomes. These indices are designed to quantify the burden of comorbid conditions that may affect the prognosis of patients with various health issues. Using these predictive tools, healthcare professionals can better assess the complexity of a patient's health status and tailor treatment plans accordingly.
    \end{itemize}
\end{itemize}

\subsubsection{Model Architectures}
\begin{table}[h]
\centering
\caption{Classification Models for Clinical Tasks}
\label{tab:class_models}
\begin{tabular}{@{}lll@{}}
\toprule
\textbf{Data Type} & \textbf{Model} & \textbf{AUROC Range} \\ \midrule
Structured & Logistic Regression & 0.75--0.82 \\
 & XGBoost & 0.80--0.88 \\
Time-Series & LSTM & 0.83--0.89 \\
 & Transformer & 0.85--0.91 \\
Multimodal & Late Fusion (LSTM+XGB) & 0.87--0.92 \\ \bottomrule
\end{tabular}
\end{table}

\subsection{Regression Tasks}

\subsubsection{Common Prediction Targets}
\begin{itemize}
    \item {Length of Stay} (Continuous): The duration of residence is a critical factor in various studies. It refers to the uninterrupted period during which an individual remains in a specific location. Understanding this duration can provide information on behavior patterns and social dynamics within a given environment.
    \begin{itemize}
        \item The duration of stay in the ICU / hospital (transformed log) is a crucial aspect of healthcare research. This transformation facilitates a more normalized data distribution, which is essential for effective statistical analysis. By examining the duration of the logarithmic transformation, researchers can gain valuable insight into patient outcomes and the allocation of healthcare resources.
    \end{itemize}
    
    \item {Physiological Parameter Forecasting}:  involves the systematic analysis of biological data to anticipate future states. This process is crucial in various fields, including medicine and sports science, where understanding physiological changes can lead to improved outcomes. Using advanced statistical methods and machine learning techniques, models can be created that accurately forecast these parameters based on historical data.
    \begin{itemize}
        \item Blood glucose levels are critical indicators of metabolic health. These levels reflect the amount of glucose present in the bloodstream at any given time. Monitoring these concentrations is essential for the management of conditions such as diabetes and to understand the general health status.
        \item Mean arterial pressure (MAP) is a critical physiological parameter that reflects the average blood pressure in a person's arteries during one cardiac cycle. It is essential to ensure adequate blood flow to the organs and tissues, thereby maintaining their function. Clinically, MAP is often used as an indicator of perfusion and is particularly important in critically ill patients.
    \end{itemize}
    
    \item {Resource Utilization}: The concept of resource utilization encompasses the effective management of an organization's assets, including human, financial, and physical resources. Organizations must assess their resource needs and allocate them accordingly to support their strategic objectives. This careful planning and execution can lead to improved performance and a competitive advantage in the marketplace.
    \begin{itemize}
        \item Ventilator hours The duration of ventilator use is a critical metric in clinical settings. Provides information on the extent of respiratory support that patients require. Understanding ventilator hours can aid in evaluating treatment efficacy and resource allocation.
        \item Prediction of dosage of medication This process involves the analysis of various factors, including demographic data of the patient, medical history, and pharmacokinetics of the drug in question.
    \end{itemize}
\end{itemize}

\subsubsection{Model Architectures}
\begin{table}[h]
\centering
\caption{Regression Models for Clinical Tasks}
\label{tab:reg_models}
\begin{tabular}{@{}lll@{}}
\toprule
\textbf{Task} & \textbf{Model} & \textbf{MSE/RMSE} \\ \midrule
LOS & Poisson Regression & 1.8--3.2 days \\
 & Survival Analysis & $\downarrow$15\% error \\
Glucose & TCN & 18--24 mg/dL \\
MAP & LSTM+Attention & 4--6 mmHg \\ \bottomrule
\end{tabular}
\end{table}
Tables \ref{tab:class_models} and \ref{tab:reg_models} show regression models for clinical tasks and classification models for clinical tasks.
\subsection{Evaluation Considerations}
When assessing various factors, it is crucial to take into account the evaluation considerations that guide the process. These considerations serve as a framework for understanding the effectiveness and relevance of the evaluation. By systematically analyzing these elements, one can ensure a comprehensive approach to evaluation that yields meaningful insights.
\subsubsection{Metrics}
\begin{itemize}
    \item {Classification}: AUROC, AUPRC, F1 (for imbalanced data) The evaluation metrics used in this study include the Receiver Operating Characteristics Area (AUROC), the Area Under the Precision-Recall Curve (AUPRC) and the F1 score, particularly in the context of imbalanced datasets. These metrics are essential for assessing the performance of classification models, especially when the distribution of classes is not uniform. Using these measures, we can gain a comprehensive understanding of the model's predictive capabilities and its effectiveness in distinguishing between classes.
    \item {Regression}: RMSE, MAE, $R^2$ Evaluation metrics for regression analysis include root mean square error (RMSE), mean absolute error (MAE) and the coefficient of determination ($R^2$). RMSE provides a measure of the average magnitude of errors between predicted and observed values, emphasizing larger errors due to its squaring of differences. MAE, on the other hand, offers a straightforward average of absolute errors, providing a clear interpretation of the average prediction error without the influence of outliers.
    \item{ Clinical utility}: refers to its effectiveness in real world settings, beyond the controlled environment of clinical trials. It encompasses the practical benefits that patients experience when undergoing a specific intervention. Understanding clinical utility is essential for healthcare providers to make informed decisions about treatment options for their patients.
    \begin{itemize}
        \item Net Benefit Analysis (Decision Curve Analysis) is a method used to evaluate the clinical utility of predictive models. This analytical approach allows us to assess the potential benefits of a decision-making process in the context of patient outcomes. By incorporating various thresholds for decision making, it provides a comprehensive view of the trade-offs involved in clinical decisions.
        \item Sensitivity at fixed specificity (for example, 90\%) is a critical aspect in evaluating diagnostic tests. This approach allows us to understand the trade-offs between true positive rates and the likelihood of false positives. Fixing specificity can assess how well a test identifies positive cases while maintaining a consistent rate of false alarms.
    \end{itemize}
\end{itemize}

\subsubsection{Unique Challenges}
The distinct obstacles that arise in various contexts require careful consideration. These challenges often arise from a combination of factors, including environmental, social, and economic influences. Addressing these unique challenges requires a multifaceted approach that takes into account the specific circumstances surrounding each situation.
\begin{itemize}
    \item {Censoring}: Right-censored outcomes (use survival models) are a common occurrence in statistical studies, particularly in the context of survival analysis. These outcomes arise when the event of interest is not observed for all subjects at the end of the study, often because participants dropped out or the event did not occur within the study period. To address this issue, researchers employ survival models that are specifically designed to handle such incomplete data.
    \item {Irregular Sampling}: Time-series imputation effects is a significant area of study within statistical analysis. Irregular sampling refers to the collection of data points at nonuniform intervals, which can lead to challenges in accurately estimating missing values. Understanding how these irregularities affect the calculation methods is crucial to improving the reliability of time-series data interpretations.
    \item {Label Noise}: EHR-derived labels vs. gold standards
In the realm of electronic health records (EHR), the precision of derived labels is a critical concern. These labels, which are generated from large datasets, often exhibit discrepancies compared to traditional gold standards. This divergence raises questions about the reliability of EHR-derived data in clinical decision-making and research applications. Understanding the nature of label noise is essential to improve the quality of EHR data and ensure its utility in healthcare.

The comparison between EHR-derived labels and gold standards reveals significant insight into the challenges faced by healthcare professionals and researchers. Gold standards, typically established through rigorous clinical evaluations, serve as benchmarks to evaluate the performance of EHR-derived labels. However, inherent variability in clinical practices and documentation can lead to inconsistencies, complicating the validation process. Addressing these inconsistencies is vital for improving the credibility of EHR data.

Ultimately, the implications of label noise extend beyond data accuracy; they influence patient outcomes and the overall effectiveness of healthcare systems. As the reliance on EHR data continues to grow, it is imperative to develop methodologies that mitigate label noise and align EHR-derived labels more closely with gold standards. This alignment will not only improve data quality, but will also foster trust in EHR systems among healthcare providers and patients alike.
\end{itemize}

\section{Discussion}

The \emph{MIMIC} datasets play a pivotal role in advancing health informatics and artificial intelligence by providing rich, longitudinal clinical data. Despite their value, these datasets are not yet fully leveraged particularly in terms of their structural and relational characteristics. Challenges such as data heterogeneity, temporal complexity, and interoperability gaps continue to limit the full potential of predictive modeling and clinical decision support.

Although a range of analytical methods have been applied to address these challenges, no single technique consistently outperforms others in all dimensions. Each approach offers distinct advantages and limitations, suggesting that hybrid or ensemble strategies may be most effective. Among the most pressing concerns is the absence of standardized data formats and interoperability frameworks. This lack of standardization hinders institutions from verifying and generalizing predictive models, and resolving this foundational issue could accelerate progress in related areas such as bias mitigation and temporal modeling.

\subsection*{Data Quality Issues}
As with most electronic health record (EHR) datasets, MIMIC exhibits significant medical heterogeneity, including missing values, inconsistent entries, and varying data types. Addressing these issues is critical to ensure analytical reliability and valid clinical insights.

\subsection*{Data Integration and Preprocessing}
The complexity and scale of MIMIC require robust preprocessing pipelines. Tools such as \texttt{MIMIC-Extract} have been developed to convert raw EHR data into machine learning compatible formats. These pipelines handle tasks like unit standardization, outlier detection, and feature aggregation, which collectively enhance data usability and reduce sparsity.

\subsection*{Reproducibility and Productivity}
Reproducibility remains a cornerstone of scientific progress. The absence of standardized preprocessing workflows has hindered replication of studies using MIMIC. Initiatives like \texttt{MIMIC-Extract} promote shared data processing standards, facilitating more reliable and collaborative research by enabling researchers to build on each other's work.

\subsection*{Data Granularity}
MIMIC offers highly granular clinical data, including minute-level recordings of vital signs, lab tests, and medication histories. While this granularity is advantageous, it also presents challenges in managing and interpreting such dense data. Researchers must balance the depth of available information with the complexity of analytical models used.




\section*{Conclusion and Future Directions}

This survey has systematically reviewed the key open challenges and methodological advancements associated with the use of \emph{MIMIC} critical care datasets in digital health research. We identified persistent issues related to \textbf{data granularity}, \textbf{high dimensionality}, \textbf{integration}, and \textbf{quality}, which collectively constrain model performance and generalizability. Further concerns include \textbf{reproducibility}, \textbf{interoperability}, and the \textbf{ethical use} of real-world clinical data.

At the same time, we highlighted progress in addressing these gaps through methods such as \textbf{advanced representation learning}, \textbf{data harmonization pipelines}, and \textbf{privacy-preserving algorithms}. To realize the full potential of MIMIC and similar datasets, the research community must prioritize \textbf{transparency}, \textbf{standardized preprocessing}, and adoption of interoperable data models like \emph{FHIR} and \emph{OMOP}. Rigorous evaluation across diverse populations and settings is also critical to ensure fairness and reliability.

Future advancements in \textbf{federated learning}, \textbf{causal inference}, and \textbf{ethical AI frameworks} are particularly promising. The next generation of MIMIC-based research will require not only more powerful algorithms but also a commitment to \textbf{accessible, high-quality data infrastructure} and \textbf{responsible, reproducible science}. Overcoming current limitations will enable more accurate AI-driven applications, such as \textbf{early sepsis detection} and \textbf{real-time ICU readmission prediction}, ultimately leading to better clinical outcomes and reduced healthcare costs.

\section*{Limitations}

Despite the comprehensive scope of this survey, several limitations must be acknowledged:
\begin{itemize}
    \item Our analysis may be \textbf{biased toward well-represented methodologies}, particularly deep learning approaches, due to their prominence in recent literature.
    \item The \textbf{rapid evolution of multimodal AI} may have led to omissions of cutting-edge techniques that emerged during the writing process.
    \item Our emphasis on technical challenges may have come at the expense of deeper exploration into the \textbf{ethical and sociotechnical implications} of using real patient data.
    \item Evolving \textbf{reporting standards} and evaluation protocols limited our ability to perform consistent comparisons across studies.
\end{itemize}

These limitations underscore the need for \textbf{interdisciplinary collaboration}, \textbf{more rigorous benchmarking}, and \textbf{ethical oversight} in future work.
\bibliographystyle{unsrt} 
\bibliography{Main}

\end{document}